\documentclass[runningheads]{llncs}

\usepackage{eccv}

\usepackage{eccvabbrv}

\usepackage{graphicx}
\usepackage{booktabs}

\usepackage[accsupp]{axessibility}  % Improves PDF readability for those with disabilities.

\usepackage[hidelinks]{hyperref}

\usepackage{orcidlink}

\begin{document}

% ---------------------------------------------------------------
% TODO REVIEW: Replace with your title
\title{Disentangled Clothed Avatar Generation from Text Descriptions} 

% TODO REVIEW: If the paper title is too long for the running head, you can set
% an abbreviated paper title here. If not, comment out.
\titlerunning{Disentangled Clothed Avatar Generation from Text Descriptions}
% TODO FINAL: Replace with your author list. 
% Include the authors' OCRID for the camera-ready version, if at all possible.
\author{
Jionghao Wang\inst{1, 2}\orcidlink{0009-0002-9683-8547}$^\dag$ \and
Yuan Liu\inst{3}\orcidlink{0000-0003-2933-5667}$^\dag$ \and
Zhiyang Dou\inst{3, 4}\orcidlink{0000-0003-0186-8269} \and \\
Zhengming Yu\inst{1}\orcidlink{0009-0003-0553-8125} \and
Yongqing Liang\inst{1}\orcidlink{0000-0002-7282-0476} \and
Cheng Lin\inst{3}\orcidlink{0000-0002-3335-6623} \and 
Rong Xie\inst{2}\orcidlink{0000-0002-8261-5337} \and \\
Li Song\inst{2}\orcidlink{0000-0002-7124-5182}$^\ddag$ \and
Xin Li\inst{1}\orcidlink{0000-0002-0144-9489} \and
Wenping Wang\inst{1}\orcidlink{0000-0002-2284-3952}$^\ddag$
}

\footnotetext[1]{$^\dag$ Co-first authors.}
\footnotetext[2]{$^\ddag$ Corresponding authors.}

% TODO FINAL: Replace with an abbreviated list of authors.
\authorrunning{J.~Wang et al.}
% First names are abbreviated in the running head.
% If there are more than two authors, 'et al.' is used.

% TODO FINAL: Replace with your institution list.
\institute{
Texas A\&M University \and
Shanghai Jiao Tong University \and
The University of Hong Kong \and
TransGP
}

\maketitle

\newcommand{\teaserCaption}{
% A demonstration of our SO-SMPL disentangled representations and its application in animation. Disentangled human body \& clothing assets generated by our method could be applied directly in physical simulation environment to obtain realistic human motions and clothes animations.
Our method generates high-quality separated human body and clothes meshes from text prompts. Kinematics or simulation motions can drive the disentangled avatar representations to achieve photorealistic animations.
}
\begin{minipage}{0.9\textwidth}
\centering 
    \includegraphics[width=\linewidth]{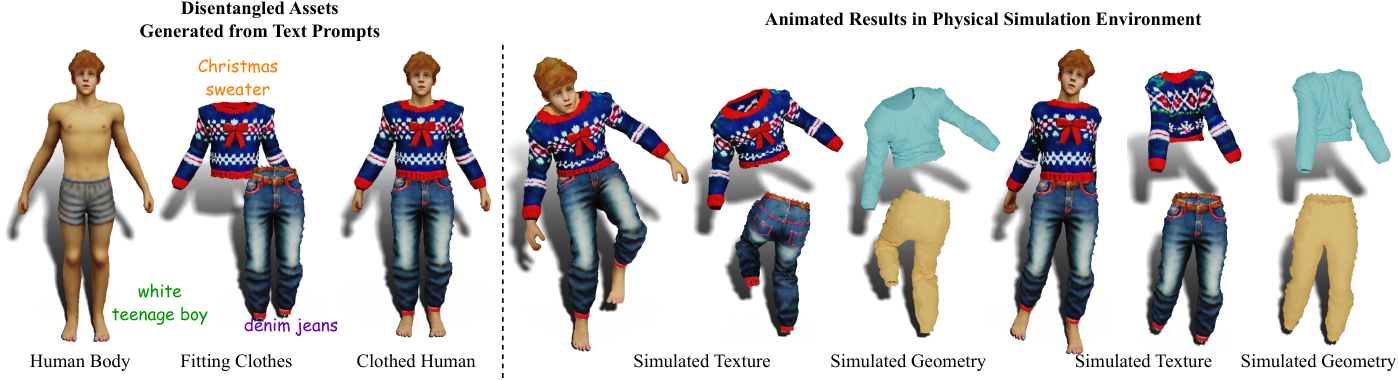}
    \captionof{figure}{\teaserCaption}
    \label{fig:teaser}
\end{minipage}

% \begin{abstract}
%   The abstract should concisely summarize the contents of the paper. 
%   While there is no fixed length restriction for the abstract, it is recommended to limit your abstract to approximately 150 words.
%   Please include keywords as in the example below. 
%   This is required for papers in LNCS proceedings.
%   \keywords{First keyword \and Second keyword \and Third keyword}
% \end{abstract}
\begin{abstract}
In this paper, we introduce a novel text-to-avatar generation method that separately generates the human body and the clothes and allows high-quality animation on the generated avatar.
While recent advancements in text-to-avatar generation have yielded diverse human avatars from text prompts, these methods typically combine all elements—clothes, hair, and body—into a single 3D representation. Such an entangled approach poses challenges for downstream tasks like editing or animation.
To overcome these limitations, we propose a novel disentangled 3D avatar representation named Sequentially Offset-SMPL (SO-SMPL), building upon the SMPL model. SO-SMPL represents the human body and clothes with two separate meshes but associates them with offsets to ensure the physical alignment between the body and the clothes. Then, we design a Score Distillation Sampling (SDS)-based distillation framework to generate the proposed SO-SMPL representation from text prompts. 
Our approach not only achieves higher texture and geometry quality and better semantic alignment with text prompts, but also significantly improves the visual quality of character animation, virtual try-on, and avatar editing. 
% We further show that our method also allows for generating complex garments, such as multi-layered clothes and skirts.
Project page: \href{https://shanemankiw.github.io/SO-SMPL/}{this link}.
\end{abstract} 

\section{Introduction}
\label{sec:intro}
Human avatars play a significant role in visual storytelling, conveying narratives through actions, expressions, and interactions with other elements in the animated world, thereby creating a dynamic and engaging experience for the audience. Avatars can be designed to represent diverse characters, including different genders, races, ethnicities, and abilities, promoting inclusivity and challenging stereotypes in the media. However, creating high-quality avatars is costly and labor-intensive, requiring skilled 3D modelers. To achieve photorealistic, animatable, and customizable avatars, it is imperative to employ a disentangled representation consisting of distinct components, such as body and clothing elements. This approach enhances customization and reusability, allowing seamless alterations to clothing while maintaining the same underlying body structure. Additionally, separate components contribute to heightened realism in animation, as different parts exhibit distinct motion characteristics. For example, human body movements can be animated using forward kinematics, while cloth motion can be generated through physical simulation.

Despite evident benefits, most existing research~\cite{huang2023tech, xiu2023econ} focuses on reconstructing a singular avatar mesh from videos or images~\cite{yu2023monohuman, qian20233dgs}, neglecting the potential for disentangled representations. Only a few studies, like DELTA~\cite{feng2023learning} and SCARF~\cite{feng2022capturing}, explore disentangled avatar reconstruction from videos but do not facilitate the arbitrary creation of such representations. Thus, a comprehensive method for creating disentangled human avatars remains an unresolved challenge. Extending text-to-avatar pipelines to generate disentangled avatars involves not only independently generating body and clothing components but also ensuring their precise alignment for visually appealing results. To address this, we introduce Sequentially-Offset-SMPL (SO-SMPL), based on SMPL-X~\cite{SMPL2015} for characterizing coarse human shapes. SO-SMPL adds trainable offsets and appearance features to represent the unclothed human body, inspired by TADA~\cite{liao2023tada}, and further refines these to learn clothing specifics. This sequential approach ensures perfect alignment and separation of body and clothing meshes.

% results
Leveraging the power of stable diffusion with SDS losses~\cite{poole2022dreamfusion}, we introduce a novel approach to generate our SO-SMPL representation from text prompts. Our method achieves disentangled avatar generation by first creating the unclothed human body and subsequently distilling the clothing components based on the underlying body structure. Experiments show that our method produces high-quality human avatars with better geometry details and appearances than baseline methods~\cite{liao2023tada,hong2022avatarclip}. The disentangled nature of our SO-SMPL representation unlocks strong customization capabilities, allowing for the seamless integration of various clothing options on the same human body. Furthermore, the avatars generated by our state-of-the-art method exhibit exceptionally realistic animation results, achieved by simulating distinct motion characteristics for both clothing and human body components. Moreover, by generating another layer of offsets, we can generate multiple-layered clothes, like a suit on a T-shirt, and each layer has a separate mesh.
% Our pioneering approach sets a new benchmark in avatar generation, opening up a world of possibilities for enhanced personalization and captivating animation experiences.

\section{Related work}
\label{sec:literature}

\noindent \textbf{Distilling diffusion for 3D tasks}.
Many recent works have utilized diffusion models to generate 3D objects or scenes. Some works~\cite{wang2023rodin, bautista2022gaudi, jun2023shap, nichol2022point, qian2023magic123, karnewar2023holodiffusion, liu2023meshdiffusion} trained 3D diffusion models to directly generate 3D representations such as point clouds, meshes, radiance fields\cite{mildenhall2021nerf}, SDFs~\cite{wang2021neus}, UDFs~\cite{yu2023surf}, DMTets\cite{shen2021deep} or tri-planes~\cite{chan2022efficient}. Others designed diffusion models to generate 3D-aware 2D images or textures\cite{shi2023mvdream, liu2023syncdreamer, long2023wonder3d, liu2023zero, wang2023360, richardson2023texture, cao2023texfusion, guo2023decorate3d, michel2022text2mesh, chen2023text2tex}. Both lines of works rely on 3D data\cite{deitke2023objaverse, deitke2023objaversexl, downs2022google} or multi-view images\cite{reizenstein2021common, yang2023synbody, yu2023mvimgnet} to train.

With the success of 2D text-to-image diffusion models~\cite{ho2020denoising, rombach2021highresolution, dhariwal2021diffusion}, recent works~\cite{poole2022dreamfusion, wang2023score, armandpour2023re, chen2023it3d, lin2023magic3d, wang2023prolificdreamer, seo2023ditto, seo2023let, chen2023fantasia3d, tsalicoglou2023textmesh, huang2023dreamtime, watson2022novel, wu2023hd, yu2023points, zhu2023hifa} propose to distill 3D representations from 2D diffusion models to circumvent the lack of 3D data. Some of them designed different 3D representations~\cite{lin2023magic3d, chen2023fantasia3d, tsalicoglou2023textmesh} for generation, while others explored the guidance schemes of diffusion models~\cite{poole2022dreamfusion, wang2023score, wang2023prolificdreamer}.

\noindent \textbf{3D Human Avatar generation}.
Previous works on 3D avatar generation~\cite{chen2022gdna, noguchi2021neural, grigorev2021stylepeople, zhang20223d, hong2022eva3d, noguchi2022unsupervised, bergman2022gnarf, zhang2022avatargen} could generate diverse human geometries or textured avatars, but the generation process could not be controlled by text descriptions. With the help of large language models like CLIP~\cite{radford2021learning} and text-to-image diffusion models~\cite{ho2020denoising, rombach2021highresolution}, recent works~\cite{cao2023dreamavatar, huang2023humannorm, zhang2023getavatar, youwangtext, kim2023chupa} were able to generate high-fidelity avatars with prompts or with both image and prompt~\cite{weng2023zeroavatar, huang2023tech, peng2024charactergen}. Other works have attempted to generate human avatars by painting the given human meshes with generated textures~\cite{svitov2023dinar, yu2023painthuman}. However, the generated avatars could not be animated and utilized in CG software. ~\cite{liao2023tada, hong2022avatarclip, kolotouros2023dreamhuman, anon2023avatarstudio, zhang2023avatarverse, huang2023dreamwaltz, liu2023humangaussian, xu2023seeavatar} could produce animatable avatars, but they ignored garment-human interactions, leading to unsatisfactory animation quality.

\noindent \textbf{Disentangled representations}.
% EVA3D: COMPOSITIONAL 3D HUMAN GENERATION FROM 2D IMAGE COLLECTIONS
Reconstructing the human body and clothes as separate geometries has long been studied in the field of computer vision~\cite{corona2021smplicit, jiang2020bcnet, feng2022capturing, feng2023learning, Patel_2020_CVPR, liao2024senchandlingselfcollisionneural}, but generating distinct body and clothes representations has barely been explored. Previous works have made attempts to generate 3D avatars as a combination of separate components~\cite{hong2022eva3d} or layers~\cite{xu2023efficient}, but they did not explicitly disentangle human body and accessories, making an individual component or layer not physically meaningful. A more recent work~\cite{hu2023humanliff} trained a diffusion-based human generation model and distinctly represented the human body and different types of clothes as separate layers, but it did not have semantic controllability and suffers from over-smoothing textures. Another effort~\cite{wang2023humancoser} proposed to use an SDS-based pipeline to generate a NeRF-based clothing layer separated from the human body, but could not be utilized in physical simulations due to its implicit representation. A concurrent work TECA~\cite{zhang2023text} explored disentangling hair and head ornament from text and images while our work focuses on decomposing the avatar into a human body and clothes. Another recent work AvatarFusion~\cite{huang2023avatarfusion} also separates clothes from human bodies. However, AvatarFusion adopts a post-processing strategy based on the outputs of AvatarClip~\cite{hong2022avatarclip}, leading to low-quality results. Our method is natively designed for disentangled generation, which produces much better quality.

% \john{Do we need to itemize our contributions? }

\section{Preliminaries}

\noindent \textbf{Score distillation sampling}~\cite{poole2022dreamfusion} is proposed to optimize a 3D representation through a frozen pre-trained large 2D diffusion model. 
% It computes a ``gradient'' for optimizing a neural field towards the direction indicated by the diffusion model. 
Specifically, to optimize a given 3D representation with parameters $\theta$ using a diffusion model with parameters $\phi$, SDS loss is computed as:
\begin{equation}
    \nabla_{\theta} \mathcal{L}_{SDS} = E_{t,\epsilon} \left[(\hat{\epsilon}_{\phi}(x_t; \mathbf{y}, t) - \epsilon) \frac{\partial x}{\partial \theta} \right]
\end{equation}
where $t$ is the time step in diffusion model, $x = g(\theta)$ is the rendered image using a differentiable renderer $g$, $x_t=x+\epsilon$ is the noise version of $x$, $\epsilon$ is a sampled gaussian noise, $\hat{\epsilon}_{\phi}(x_t; \mathbf{y}, t)$ is the predicted noise of the diffusion model, and $\mathbf{y}$ is the input text prompt.
% Note that the Jacobian term for the denoising U-Net in Stable Diffusion is circumvented directly for efficiency concerns. 

\noindent \textbf{SMPL-X}~\cite{pavlakos2019expressive} is an expressive parametric human model that could produce a human mesh with a fixed topology. 
% SMPL-X contains $N=10,475$ vertices and 20908 faces. 
SMPL-X is conditioned by a given input shape parameter $\beta$, pose parameter $\theta$, and expression parameters $\psi$, and follows a vertex-based linear blend skinning(LBS) based on skeleton $J(\beta)$ and skinning weights $W$ for pose transforming and animation. Specifically, SMPL-X computes a posed human mesh $\mathbf{K}(\beta, \theta, \psi)$ as:
\begin{align}
\begin{split}
    \mathbf{K}(\beta, \theta, \psi) &= \mathcal{W}(\mathbf{T}(\beta, \theta, \psi), J(\beta), \theta, W) \\
    \mathbf{T}(\beta, \theta, \psi) &= T + B_s(\beta) + B_e(\psi) + B_p(\theta)
\label{equ:smplx}
\end{split}
\end{align}
%\ly{explain the symbols here.}
where an T-pose body mesh $\mathbf{T}(\beta, \theta, \psi)$ is first calculated as the combination of a mean template $T$, shape, expression and pose blend shapes $\{B_s, B_e, B_p\}$ then warped to target pose $\theta$ with LBS operation $\mathcal{W}$.

\section{Methodology}
\begin{figure*}[t]
  \centering
  %\fbox{\rule{0pt}{2in} \rule{\linewidth}{0pt}}
   \includegraphics[width=\linewidth]{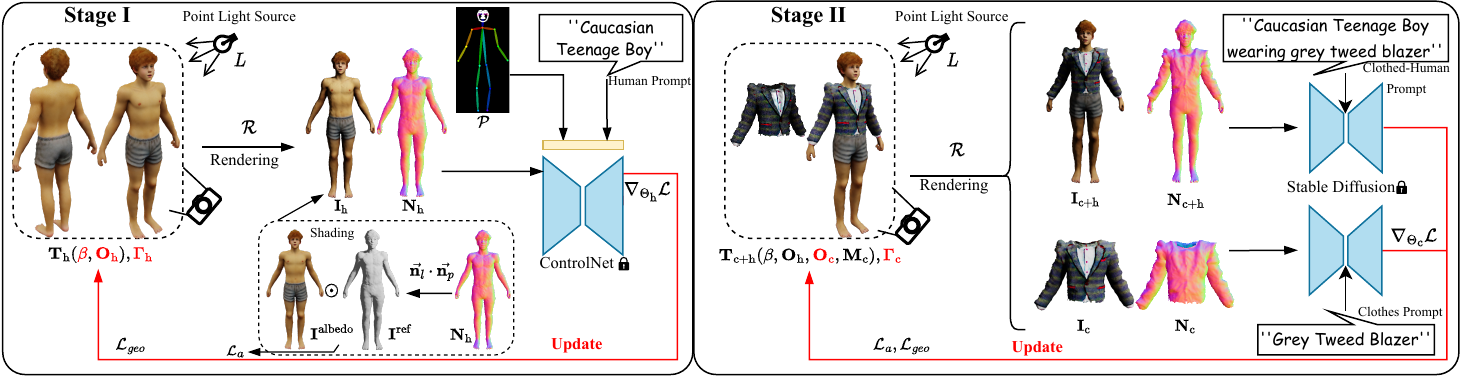}

   \caption{\textbf{An overview of our generation pipeline.} Our pipeline has two stages. In Stage I, we generate a base human body model by optimizing its shape parameter $\beta$, vertex offset $\mathbf{O}_{\text{h}}$ and albedo texture $\Gamma_{\text{h}}$. A ControlNet is utilized to compute a Score Distillation Sampling(SDS) Loss conditioned on an ``A''-pose map $\mathcal{P}$ for the rendered RGB image $\mathbf{I}_{\text{h}}$ and normal map $\mathbf{N}_{\text{h}}$. In Stage II, we freeze the human body model and optimize the clothes parameters $\mathbf{O}_{\text{c}}$ along with the albedo texture MLP $\Gamma_{\text{c}}$. The rendered RGB images and normal maps of both the clothed human and the clothes are used in computing the SDS losses. In both stages, we utilized a simple Phong shading model to render images from our SO-SMPL representations.}
   \label{fig:train_pipeline}
\end{figure*}
Fig.~\ref{fig:train_pipeline} is an overview of our disentangled avatar generation pipeline. Given text descriptions of the human avatar, our pipeline produces an animatable and high-quality unclothed avatar in the first stage. In the second stage, we create clothes on the target avatar.
% TODO:
% 1. Simplify math equations and symbols
% 2. Simplify the method section and make it clearer
% 3. Address
% Specifically, we aim to optimize the base human body geometry $\mathbf{T}_{\text{h}}$ and clothes geometry $\mathbf{T}_{\text{c}}$, along with their respective textures $\Gamma_{\text{h}}$ and $\Gamma_{\text{c}}$, by applying SDS gradients through normal maps and color images rendered by the renderer $\mathcal{R}$.
% \begin{align}
% \begin{split}
%     \mathbf{N}_{\text{h}} = \mathcal{R}(\mathbf{M}_{\text{h}}, \pi), \mathbf{I}_{\text{h}} = \mathcal{R}(\mathbf{M}_{\text{h}}, \Gamma_{\text{h}}, \pi)  \\
%     \{\mathbf{N}_{\text{c}},\mathbf{N}_{\text{c+h}}\} = \mathcal{R}(\mathbf{M}_{\text{c}}, \pi), \{\mathbf{I}_{\text{c}},\mathbf{I}_{\text{c+h}}\}  = \mathcal{R}(\mathbf{M}_{\text{c}}, \Gamma_{\text{c}}, \pi)
% \end{split}
% \label{equ:render}
% \end{align}
% where $\pi$ is a specific camera pose and $\{\mathbf{N}_{\text{h}}$, $\mathbf{I}_{\text{h}}\}$, $\{\mathbf{N}_{\text{c}}, \mathbf{I}_{\text{c}}\}$, $\{\mathbf{N}_{\text{c+h}}$, $\mathbf{I}_{\text{c+h}}\}$ are rendered normal maps $\mathbf{N}$ and color images $\mathbf{I}$ for human body, clothes, clothed human, respectively.
In the following, we introduce our disentangled representation, called SO-SMPL, of the human body and the clothes.

%In Sec. \ref{sec:geometry}, we will introduce our geometry representations ${\mathbf{M}_{\text{h}}, \mathbf{M}_{\text{c}}}$; in Sec. \ref{sec:texture}, our texture modeling ${\Gamma_{\text{h}}, \Gamma_{\text{c}}}$ will be illustrated.
%As could be seen from Fig. , our method firstly generates a base avatar based on 

\subsection{SO-SMPL Representation}
\label{sec:geometry}
%\ly{double-check: human model or human body model?}
% In order to generate clothes and human body models separately, we opt to model them with distinct geometry representations. For the generation of a human body, we utilize the parameter space of the shape parameter $\beta$ mentioned in Eq.~\ref{equ:smplx} to model the variety of coarse body shapes and add vertex-wise offsets to represent fine geometry details. 
% As for clothes, we propose to use extra vertex-wise offsets and masks on top of the base human body to represent a single-layer clothes mesh. Both human body and clothes are generated in  fixed ``A"-pose.

\begin{figure}[t]
  \centering
  %\fbox{\rule{0pt}{2in} \rule{\linewidth}{0pt}}
   \includegraphics[width=0.8\linewidth]{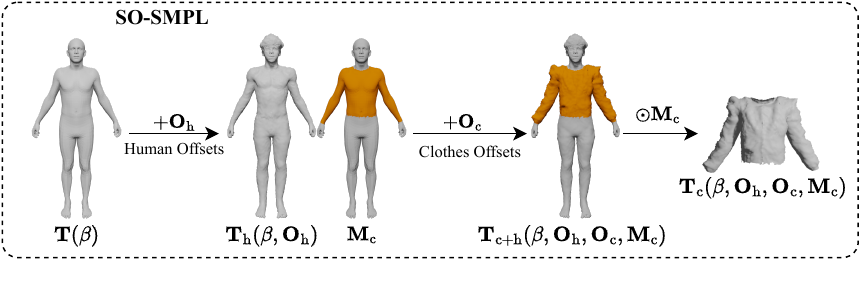}

   \caption{\textbf{Illustration of our SO-SMPL representation.} Two vertex-wise offsets, namely the human offset $\mathbf{O}_{\text{h}}$ and the clothes offset $\mathbf{O}_{\text{c}}$ are sequentially added in order to the SMPL-X body mesh $\mathbf{T}(\beta)$ to obtain the human body mesh $\mathbf{T}_{\text{h}}(\beta, \mathbf{O}_{\text{h}})$ and clothed human mesh $\mathbf{T}_{\text{c+h}}(\beta, \mathbf{O}_{\text{h}}, \mathbf{O}_{\text{c}}, \mathbf{M}_{\text{c}})$, where a vertex mask $\mathbf{M}_{\text{c}}$ is calculated to determine the clothing region. Finally, we mask the clothed human mesh with $\mathbf{M}_{\text{c}}$ and obtain the separated clothes mesh $\mathbf{T}_{\text{c}}(\beta,  \mathbf{O}_{\text{h}},\mathbf{O}_{\text{c}},\mathbf{M}_{\text{c}})$.}
   \label{fig:mask}
\end{figure}

\noindent \textbf{Human body geometry.} As shown in Fig.~\ref{fig:mask}, we represent the human body mesh with a \textit{densified} SMPL-X parametric model introduced in Eq.~\ref{equ:smplx} and we add learnable vertex-wise offsets $\mathbf{O}_{\text{h}}$ to represent details on the body, similar to recent works~\cite{liao2023tada}. We fix the pose of the human body mesh to an ``A'' pose $\theta$ and use a default expression parameter $\psi$ during the text-to-avatar generation process so that $\theta$ and $\psi$ are omitted for simplicity. Specifically, the human body geometry $\mathbf{T}_{\text{h}}$ is
\begin{align}
\begin{split}
    % \mathbf{K}_{\text{h}}(\beta, \theta, \psi, \mathbf{O}_{\text{h}}) &= \mathcal{W}(\mathbf{T}_{\text{h}}(\beta, \theta, \psi,  \mathbf{O}_{\text{h}}), J(\beta), \theta, W)  \\
    \mathbf{T}_{\text{h}}(\beta, \mathbf{O}_{\text{h}}) &= \mathbf{T}(\beta) + \mathbf{O}_{\text{h}},
\end{split}
\label{equ:smplplus}
\end{align}
where $\mathbf{O}_{\text{h}}$ is trainable vertex-wise offset and $\beta$ is trainable shape parameter.
% where the offset $\mathbf{O}_{\text{h}}$ for $N_s$ dense body vertices is added to the mesh after the subdivision operation $\mathcal{S}: \mathbb{R}^{N \times 3}  \rightarrow \mathbb{R}^{N_s \times 3}$.
% , while  $W \in \mathbb{R}^{N_s\times J}$ is the dense skinning weights

\noindent \textbf{Clothes geometry.}
Clothes are represented by additional vertex-wise masks $\mathbf{M}_{\text{c}}$ and offsets $\mathbf{O}_{\text{c}}$ on $\mathbf{T}_{\text{h}}$. 
% To enable the generation of diverse clothes types, we pre-defined vertex-level templates of 6 different types: long shirt, short shirt, long pants, short pants, vest, and overalls. 
The clothed human mesh $\mathbf{T}_{\text{c+h}}$ are
\begin{align}
\begin{split}
 \mathbf{T}_{\text{c+h}}(\beta, \mathbf{O}_{\text{h}},\mathbf{O}_{\text{c}},\mathbf{M}_{\text{c}}) &= \mathbf{T}_{\text{h}}(\beta, \mathbf{O}_{\text{h}}) + \mathbf{O}_{\text{c}} \odot \mathbf{M}_{\text{c}}, 
\end{split}
\label{equ:smplplusclothedhuman}
\end{align}
where $\odot$ denotes the Hadamard product between the vertex-wise offsets and masks. 
% To separate the clothes mesh while preserving the surface characteristics of the segmented clothes vertices, we re-used the topology of the human body mesh for the clothes mesh. 
We can also get the clothes geometry $\mathbf{T}_{\text{c}}$ only by masking on the clothed human mesh
\begin{align}
\begin{split}
 \mathbf{T}_{\text{c}}(\beta,  \mathbf{O}_{\text{h}},\mathbf{O}_{\text{c}},\mathbf{M}_{\text{c}})   &= \mathbf{T}_{\text{c+h}}(\beta, \mathbf{O}_{\text{h}},\mathbf{O}_{\text{c}},\mathbf{M}_{\text{c}}) \odot \mathbf{M}_{\text{c}}.
\end{split}
\label{equ:smplplusclothes}
\end{align}
To enable the generation of diverse clothes types, we use the body-part segments of SMPL-X and pre-define vertex-level mask templates of 6 different garment types, namely long shirts, short shirts, long pants, short pants, vests, and overalls. These templates are used to initialize clothes mask $\mathbf{M}_{\text{c}}$. We can also support skirts generation using the skirts template from BCNet~\cite{jiang2020bcnet}.
% During the first stage(human body generation), the SMPL-X shape parameter $\beta$, the body displacement $\mathbf{O}_{\text{h}}$ and $\xi_{\text{h}}$ that parameterize the human texture MLP $\Gamma_{\text{h}}$ are optimized, while the expression parameter $\psi$ and pose parameter $\theta$ are locked to default values. On the other hand for garment generation, only the clothes offset $\mathbf{O}_{\text{c}}$ and $\xi_{\text{c}}$ that parameterize the clothes MLP $\Gamma_{\text{c}}$ are updated through the generation process.

\noindent \textbf{Discussion}. 
% \ly{todo} 
Previous works~\cite{zhang2023text,huang2023avatarfusion} tried to represent clothes as a volume-based representation, e.g. NeRF~\cite{mildenhall2021nerf}, which is hard to be converted to a mesh. Our clothes representation is naturally a thin mesh representation, which is preferable for most graphics software and physical simulators~\cite{MarvelousDesigner, CLO3d}. Meanwhile, by representing the clothes as a displacement surface that ``grows'' upon the base human, we ensure the alignment between the human body and clothes. 

\subsection{SO-SMPL Rendering}
\label{sec:texture}

On both the human body mesh and the clothes mesh, we model appearances with a Phong shading model where the meshes are characterized by their albedo and the lighting is simply a combination of a point light and an ambient light. 
% Consistent with our disentangled geometry design, we assigned distinct learnable texture representations, namely $\Gamma_{\text{h}}$ and $\Gamma_{\text{c}}$ to model human body and clothes appearances, respectively. In both the human body and clothes mesh generation stages, the texture is optimized jointly with geometry.

% Fig.~\ref{fig:render_pipeline} depicts our disentangled render schemes for the two stages of our pipeline. As mentioned in Eq.~\ref{equ:render}, we use two texture representations $\Gamma_{\text{h}}$ and $\Gamma_{\text{c}}$ to represent the texture of the human body and the clothes, respectively. 
%\ly{this explanation is too early to show here and is not very clear.} During animation simulations, the quality of texture is essential to the realism of the results. For SDS-based 3D generative methods, the updating gradient is measured merely by the rendered image from a single view. Therefore, the lack of 3D context poses two major challenges for textures generated by SDS-guidance-based methods: inherent geometry and inherent lighting. To be specific, geometric patterns(e.g. wrinkles and bumps) and lighting patterns(shadows and highlights) are integrately learned into the albedo of the model. These inherent effects are extremely harmful and could cause major artifacts for downstream applications such as character animation and clothes simulation. To alleviate these effects, we propose to sample point light sources during training, and model albedo and textureless color produced by lighting and surface normals distinctly. 

\noindent \textbf{Albedo}.
% With a fixed vertex-face topology, one can naively use a fixed UV mapping and target to refine a fixed texture map during optimization\cite{liao2023tada}. However, the limited pattern of texture map and lack of spatial-awareness of the vertex positions could lead to unrealistic textures. Meanwhile, in the case of clothes, the produced mesh topology could be changing during training, and hence no fixed UV mapping is available. Following previous works~\cite{poole2022dreamfusion, Chen_2023_ICCV},
We use an MLP $\Gamma_{\text{i}}$ to represent the albedo of the mesh. The albedo $\mathbf{\rho}$ of a 3D point $\mathbf{x}$ on the target mesh is computed by
\begin{align}
 \mathbf{\rho}  &=  \Gamma_i(\gamma(\mathbf{x}); \xi_i),
\label{equ:mlp}
\end{align}
where $i \in \{\text{h},\text{c}\}$ means clothes or human body, $\gamma$ is positional encoding of the 3D position $\mathbf{x}$, and $\xi_i$ is the parameter of the MLP. 

\noindent \textbf{Shading model.}
\label{sec:light}
%During animation, it is crucial to disentangle lighting conditions from the intrinsic albedo; otherwise, unnatural, persistent shadows and wrinkles may emerge under varying lighting conditions, leading to a suboptimal visual experience. 
%To tackle the issue, we employed\ZY{are you using an off-the-shelf method?} a diffuse lighting model~\cite{} for simulating cloth textures, effectively preventing the incorporation of lighting conditions into the model's intrinsic albedo. This approach ensures a more visually appealing representation of the animated garments, regardless of the external lighting environment.
Given a randomly-sampled point light with a position of $\mathbf{v}$ and an intensity of $\mathbf{l}_{\text{d}}$, along with the ambient light intensity $\mathbf{l}_{\text{a}}$, the shading color $\mathbf{c}$ of a surface point $\mathbf{p}$ is computed by
\begin{align}
\begin{split}
 \mathbf{c}   &= \mathbf{\rho}_\mathbf{p} (\mathbf{l}_{\text{a}} + \max(0, \vec{\mathbf{n}}_l \cdot \vec{\mathbf{n}}_\mathbf{p}) \mathbf{l}_{\text{d}})
\end{split}
\label{equ:lambertian}
\end{align}
where $\vec{\mathbf{n}}_\mathbf{p}$ is the normal direction of the point $\mathbf{p}$, $\vec{\mathbf{n}}_l = (\mathbf{v} - \mathbf{p})/\|\mathbf{v} - \mathbf{p}\|_2$ is the light direction towards the surface vertice, and $\rho_{\mathbf{p}}$ is the albeodo on this point.

\begin{figure}[htbp]
 \centering
  \includegraphics[width=0.6\linewidth]{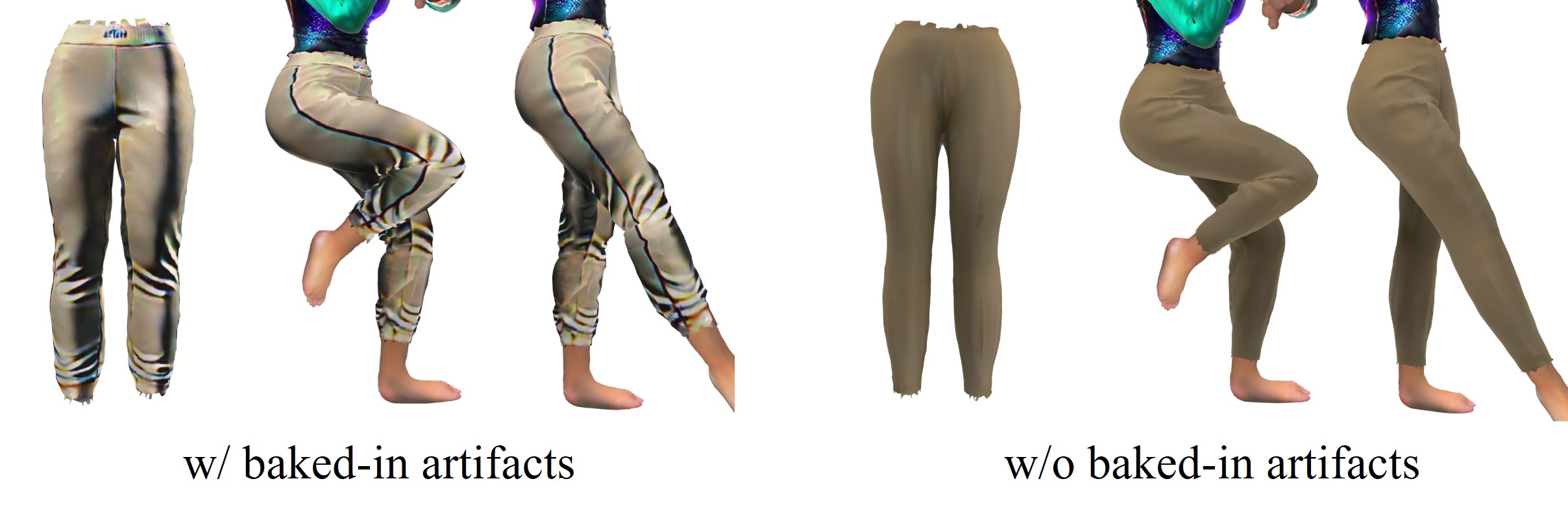}

  \caption{\textbf{An intuitive illustration of the impact of the baked-in artifacts in learned albedo.} On the left, the generated pant suffers from severe baked-in wrinkles in its texture, resulting in non-photorealistic wrinkles and shadows in animation. On the right side, our shader prevents the shadows from baking into the albedo, hence significantly improving the visual quality of animations.}
  \label{fig:baked_illustrate}
\end{figure}

\noindent \textbf{Discussion}. 
% In order to make the trained model adapt to different lighting conditions, we followed \cite{poole2022dreamfusion, lin2023magic3d} to randomly sample point light sources and compute the light reflectance during training. These known light sources are used to calculate textureless shading with respect to the rendered surface normal, combining the ambient light and Lambertian diffuse reflectance term\cite{lambert1760photometria}. 
Incorporating this shading model greatly reduces the chance that the shadows are baked into the albedo of the generated clothes, which improves the consistency between geometry and textures, as shown in Fig.~\ref{fig:baked_illustrate}. 
This shading model associates the rendering colors with the generated normals so optimizing rendering colors also changes the normal directions, which reduces the chances of getting baked-in shadows or wrinkles. 
% This enhancement can be attributed to our SO-SMPL's ability to compute normals with greater accuracy compared to previous methods. These precisely estimated normals facilitate our shading model's capacity to explicitly render shadows, particularly those resulting from wrinkles on the clothes.
% Different from \cite{poole2022dreamfusion, lin2023magic3d}, since we have a plausible initial geometry, we use this light decomposition model throughout the whole generation process. In this way, the geometric bumps on the clothings \& humans would result in corresponding shadows \& wrinkles, which could help preventing the inherent patterns and the misalignment of geometry and texture.

\noindent \textbf{Rendering clothes \& human body.}
From any given camera viewpoint, we can render the images from our SO-SMPL representation using rasterization. We first determine the corresponding 3D point and its normal from $\mathbf{T}_{\text{h}+\text{c}}$ on every pixel. Then, the shading colors are computed using Eq.~\ref{equ:lambertian}, resulting in the rendered image $\mathbf{I}_{\text{h}+\text{c}}$. In a similar fashion, we render separate images for the body ($\mathbf{I}_{\text{h}}$) and the clothes ($\mathbf{I}_{\text{c}}$), respectively. Concurrently, we also obtain the corresponding normal maps $\mathbf{N}_{\text{h}+\text{c}}$, $\mathbf{N}_{\text{h}}$, and $\mathbf{N}_{\text{c}}$. More details are included in the supplementary material.
% To render the clothed human rgb image $\mathbf{I}_{\text{c+h}}$ from separate body and clothes textures, we rasterize the SO-SMPL with its vertex-level mask $\mathbf{M}_{c}$ to obtain 2D blending mask $\mathbf{I}^{\text{mask}}$ and use it to combine the rgb values of human body $\mathbf{I}_{\text{h}}$ \& clothing $\mathbf{I}_{\text{c}}$, i.e. :
% \begin{align}
% \begin{split}
%  \mathbf{I}_{\text{c+h}}   &= \mathbf{I}_{\text{c}} \odot \mathbf{I}^{\text{mask}} + \mathbf{I}_{\text{h}} \odot (\mathbf{1} - \mathbf{I}^{\text{mask}})
% \end{split}
% \label{equ:blend}
% \end{align}
% where $\odot$ denotes the Hardmard product. Notably, self-occlusions are already considered in the rasterization process thus $\mathbf{I}^{\text{mask}}$ could be applied directly. 
% The computation of the clothes and human textures could also be calculated as $\mathbf{I}^{\text{albedo}}_{\text{h}} = \mathbf{I}^{\text{albedo}}_{\text{h}} \cdot \mathbf{I}^{\text{ref}}_{\text{h}}$ and $\mathbf{I}^{\text{albedo}}_{\text{c}} = \mathbf{I}^{\text{albedo}}_{\text{c}} \cdot \mathbf{I}^{\text{ref}}_{\text{c}}$.

\subsection{SO-SMPL Generation}

To learn the geometry and texture representations of the SO-SMPL model, we utilize SDS losses on both the rendered RGB images and the normal maps. Our approach involves a two-stage optimization process: In Stage I, we focus on generating an unclothed human body model. Following this, in Stage II, we proceed to generate the clothing, which is attached to the human body model established in Stage I.
% To ensure the generated 3D assets could produce plausible visual results in physical simulation environments, we also added albedo smoothness constraints and geometry constraints.

\noindent \textbf{Stage I: Human body generation.}
% Due to the randomness and diversity from a multi-step schedule, the predicted 2D noise from the U-Net in diffusion models~\cite{rombach2021highresolution} varies in terms of color and its positions on the image. However, an inherent advantage of SMPL-X representation is its strong prior knowledge of the human joints pose. To this end, we propose to utilize the OpenPose~\cite{cao2017realtime}-based ControlNet~\cite{zhang2023adding} that takes an extra 2D joint map as input condition that guides the predicted noise. 
For the human body generation, we adopt an OpenPose~\cite{cao2017realtime}-based ControlNet~\cite{zhang2023adding} with SDS~\cite{poole2022dreamfusion} losses to optimize a SO-SMPL representation. An ``A''-pose joint map $\mathcal{P}$ rendered from 3D skeleton joints is utilized as the condition for ControlNet to compute SDS losses. Given an input rendered image $\mathbf{I} \in \{\mathbf{I}_{\text{h}}, \mathbf{N}_{\text{h}}\}$, the gradient is computed by
%Specifically, given a camera view $\pi$, a joint map $\mathcal{P}$ is rendered based on the shape parameters $\beta, \theta, \psi$, i.e. $\mathcal{P}  = \mathcal{R}_J(\mathcal{J}(\beta, \theta, \psi), \pi)$, where $\mathcal{J}: \mathbb{R}^{|\boldsymbol{\beta}| + |\boldsymbol{\theta}| + |\boldsymbol{\psi}|} \rightarrow \mathbb{R}^{N_J \times 3}$ and $N_J$ is the number of joints. 
\begin{equation}
    \nabla_{\Theta_{\text{h}}} \mathcal{L} = E_{t,\epsilon} \left[(\hat{\epsilon}_{\phi}(\mathbf{I}_t; \mathbf{y}_{\text{h}}, \mathcal{P}, t) - \epsilon) \frac{\partial \mathbf{I}}{\partial \Theta_{\text{h}}} \right],
\end{equation}
where $\mathbf{I}_t=\mathbf{I}+\epsilon$ is a noisy version of the input image, $\Theta_{\text{h}}=\{\beta, \mathbf{O}_{\text{h}},\xi_{\text{h}}\}$ is the trainable parameters during human body generation, $\{\beta, \mathbf{O}_{\text{h}}\}$ are geometry parameters, $\xi_{\text{h}}$ is the texture parameters, $\hat{\epsilon}_{\phi}(\mathbf{I}_t; \mathbf{y}_{\text{h}}, \mathcal{P}, t)$ is the predicted noise given text embedding $\mathbf{y}_{\text{h}}$, noise step $t$ and the pose map $\mathcal{P}$. To ensure our generated human body mesh is unclothed, we add specific descriptions to the human prompt such as ``wearing tight shorts''. Additionally, to further guide the body mesh and texture away from depicting clothing, we employ negative prompts like ``loose clothes, accessories''.

\noindent \textbf{Stage II: Clothes generation.}
Similar to human body generation, we also utilize SDS losses to update clothing parameters $\Theta_{\text{c}} = \{\mathbf{O}_{\text{c}}, \xi_{\text{c}}\}$. We calculate SDS gradients through RGB and normal images of both the clothed human and the separated clothes. Given a rendered image $\mathbf{I} \in \{\mathbf{I}_{\text{c}}, \mathbf{N}_{\text{c}}, \mathbf{I}_{\text{c+h}}, \mathbf{N}_{\text{c+h}}\}$, and its corresponding text prompt embedding $\mathbf{y}_{\text{c}}$, the SDS loss is termed as:
\begin{align*}
\begin{split}
    \nabla_{\Theta_{\text{c}}} \mathcal{L} = E_{t,\epsilon} \left[(\hat{\epsilon}_{\phi}(\mathbf{I}_t; \mathbf{y}_{\text{c}}, t) - \epsilon) \frac{\partial \mathbf{I}}{\partial \Theta_{\text{c}}} \right]
\end{split}
\end{align*}

\noindent \textbf{Albedo smoothness constraint.}
To avoid geometries (e.g. wrinkles and bumps) and lighting (shadows and highlights) being baked into the albedo of the model, we employ an albedo smoothness loss following~\cite{zhang2021nerfactor, chen2022tracing} to reduce these effects. With a small perturbation $\delta$ of the input position $\mathbf{x}$, we encourage the estimated albedo from texture MLP $\Gamma_i \in \{\Gamma_{\text{h}}, \Gamma_{\text{c}}\}$ to be consistent, i.e.
\begin{equation}
\mathcal{L}_{a} = \|\Gamma_i(\mathbf{x}) - \Gamma_i(\mathbf{x} + \delta)\|_2
\end{equation}

\noindent \textbf{Geometry constraints.}
In both the human body and clothing generation, we utilize geometry constraints to ensure that the mesh is smooth and does not deviate too far from the original SMPL-X mesh. We employ a Laplacian smoothness term~\cite{kanazawa2018learning} $\mathcal{L}_{s}$ to minimize the norm of graph Laplacian. We also add an offset regularizing term $\mathcal{L}_{o} = \|\mathbf{O}_i\|_2$ for the body/clothes offset $\mathbf{O}_i \in \{\mathbf{O}_{\text{h}}, \mathbf{O}_{\text{c}}\}$. Thus, the geometry regularization loss is $\mathcal{L}_{\text{geo}} = \mathcal{L}_{s} + \mathcal{L}_{o}$. More implementation details such as camera configurations are included in the supplementary material.
% Mention negative prompting in supp. Don't need to mention it in here probably.
% \noindent \textbf{Negative Prompting Guidance}\cite{ho2022classifier}
% In order to disdinguish clothes and the unclothed human, we need to make sure the learned human does not contain any loose clothes and vice versa for the generated clothes. To tackles this issue, we opt to add addtional negative prompts to guide the optimization progress. As proposed in \cite{ho2022classifier}, a linear combination of conditional and unconditional score estimation could be used as the ``guided'' noise direction. Instead of unconditional input, we could also use a negative condition $\mathbf{y}_{\text{neg}}$ to direct the generative process ``away'' from the undesired conditions, i.e.:
% \begin{align}
%     \hat{\epsilon}_{\phi}(x_t; \mathbf{y}, t) &= (s+1)\epsilon_{\phi}(x_t; \mathbf{y}, t) - s\epsilon_{\phi}(x_t; \mathbf{y}_{\text{neg}}, t)
% \end{align}
% where $\mathbf{y}$ and $\mathbf{y}_{\text{neg}}$ are input text embeddings for positive and negative prompts, respectively, and $s$ is the guidance scale. 

% During animation, if the lighting condition is not disentangled from the inherent albedo, there would be unnatural inherent shadows and wrinkles that stays unchanged for different lighting conditions, which could cause bad visual experience.
% We used diffuse lighting model to model the clothes texture. This could avoid lighting conditions to be learned into the inherent albedo of the model. 

% 

\section{Experiments}
% We validate the proposed framework on the task of disentangled avatar generation. We first demonstrate the diversity of our generation pipeline in Sec. Then, in Sec.~\ref{sec:comp_quality} and Sec.~\ref{sec:user_study}, we evaluate and compare our methods with existing avatar generation methods in generation quality with both qualitative comparisons and user studies. We also present ablation analysis and applications in Sec.~\ref{sec:ablation} and Sec.~\ref{sec:application}, respectively.
\subsection{Generated Avatars \& Clothes}
\begin{figure*}[t]
  \centering
   \includegraphics[width=\linewidth]{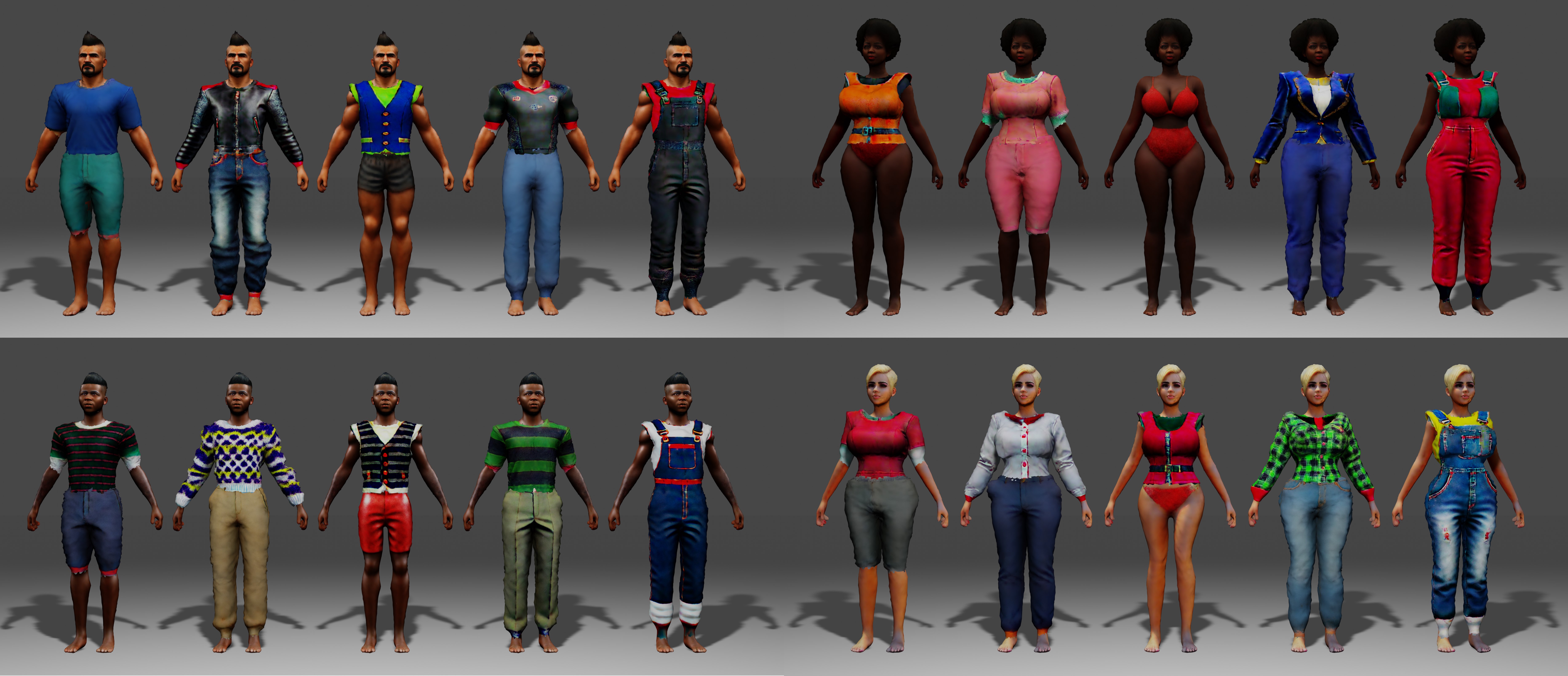}

   \caption{\textbf{A gallery of our generated clothed avatars.} Our method can generate avatars with varied ethnicities, genders, and clothes. 
   % In the figure, we demonstrated avatars generated from 4 different prompts: a strong Hispanic man with a mohawk haircut, a beautiful African-American woman, a young white teenage boy, and a thin white teenage girl. 
   % We also presented various generated clothing items that fits their body shapes. 
   %\john{The right two do not look right. Should modify the image. Also, might need to add text prompts for these clothes.}
   }
   \label{fig:comp_diverse}
\end{figure*}
 
%We generated animatable avatars of diverse ethnicities and body types. 

As demonstrated in Fig.~\ref{fig:comp_diverse}, our method generates diverse avatars and various types of clothes with detailed textures. Besides, since the clothes are generated on a base human body, they fit with the intended human naturally.  

\subsection{Quality Comparisons}
\label{sec:comp_quality}
We conduct comparisons on three aspects, static clothed-avatar quality, clothes quality, and animation quality. For the generation quality of clothed avatars, we compare with state-of-the-art text-to-3D avatar generation methods TEXture~\cite{richardson2023texture}, AvatarCLIP~\cite{hong2022avatarclip} and TADA!~\cite{liao2023tada}. As for clothes generation, we conduct comparisons with two of the best existing text-to-3D methods ProlificDreamer~\cite{wang2023prolificdreamer} and TextMesh~\cite{tsalicoglou2023textmesh}. We then compare animation results in the physical simulation environment~\cite{MarvelousDesigner} with the animatable meshes generated by TEXture~\cite{richardson2023texture} and TADA!~\cite{liao2023tada}.

\noindent \textbf{Clothed avatar quality.}
\begin{figure*}[t]
  \centering
   \includegraphics[width=\linewidth]{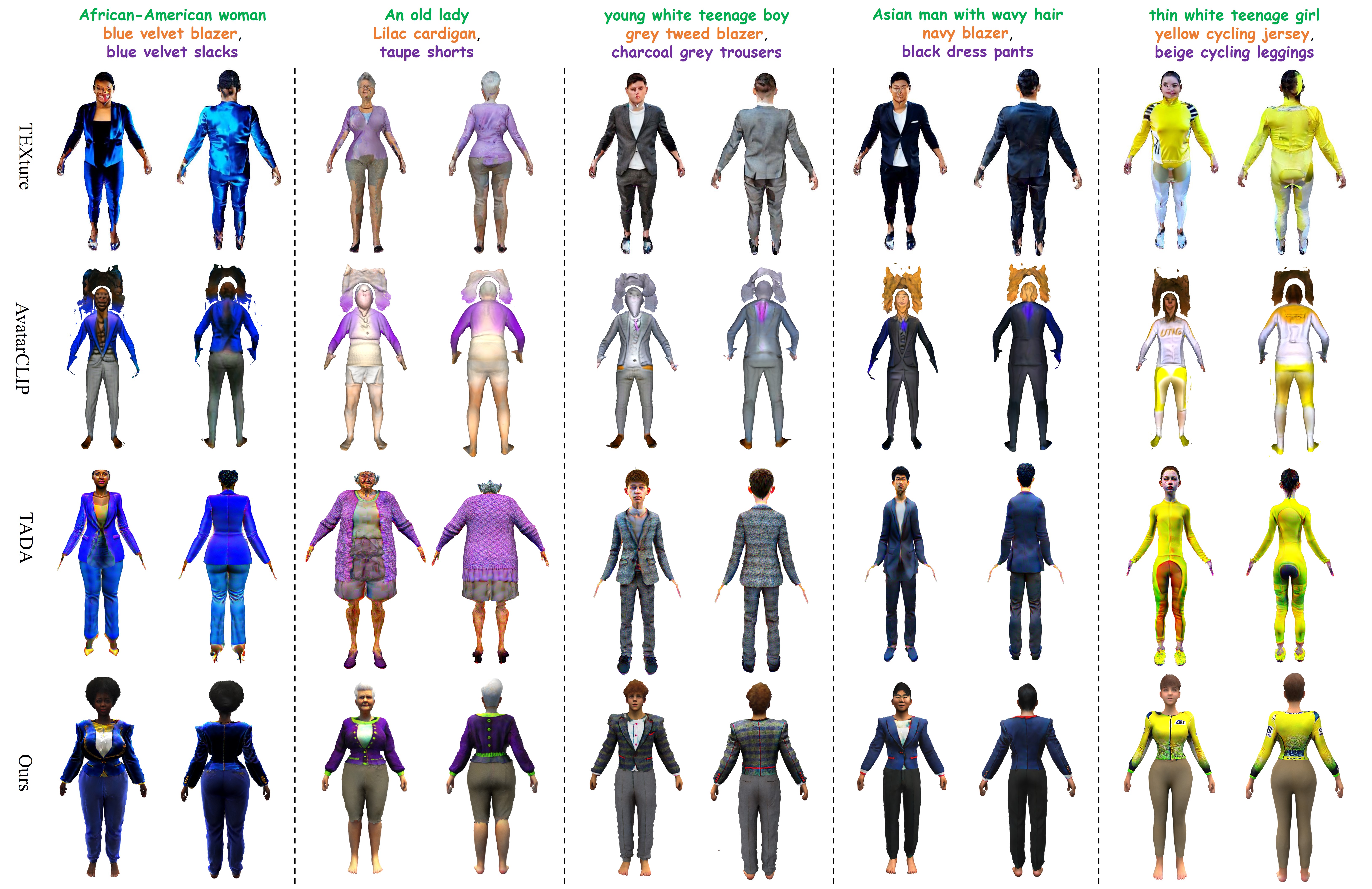}

   \caption{\textbf{Quality comparison of the generated clothed avatars}. We compare results from  TEXture~\cite{richardson2023texture}, AvatarCLIP~\cite{hong2022avatarclip}, TADA!~\cite{liao2023tada} and our method with the same prompts. 
   % As could be observed from the results, our method generates avatars with not only higher geometry and texture quality, but also more semantic consistency with the text descriptions. 
   %\john{Pick some worse examples for TADA. Maybe the old lady or somebody else.}
   }
   \label{fig:comp_static}
\end{figure*}
A visual comparison of the generated avatars is presented in Fig.~\ref{fig:comp_static}. TEXTure~\cite{richardson2023texture} takes a fixed SMPL-X mesh as input and suffers from texture inconsistency under different views. AvatarCLIP~\cite{hong2022avatarclip} often produces geometry artifacts and poor facial details. TADA!~\cite{liao2023tada} tends to generate unnatural geometric artifacts and unaligned outfits, e.g. a hull on the African-American woman's belly and the white teenage girl's yellow pants that contradicts the ``beige legging'' described by the prompt. This is because they use a simple text-to-image diffusion model and lack sufficient regularization on geometry. In comparison, using a pose-controlled diffusion model and proper regularization, our method can produce not only the most realistic human with detailed textures but also high-quality clothes that are semantically aligned with texts.

\begin{figure}[!ht]
  \centering
   \includegraphics[width=0.8 \linewidth]{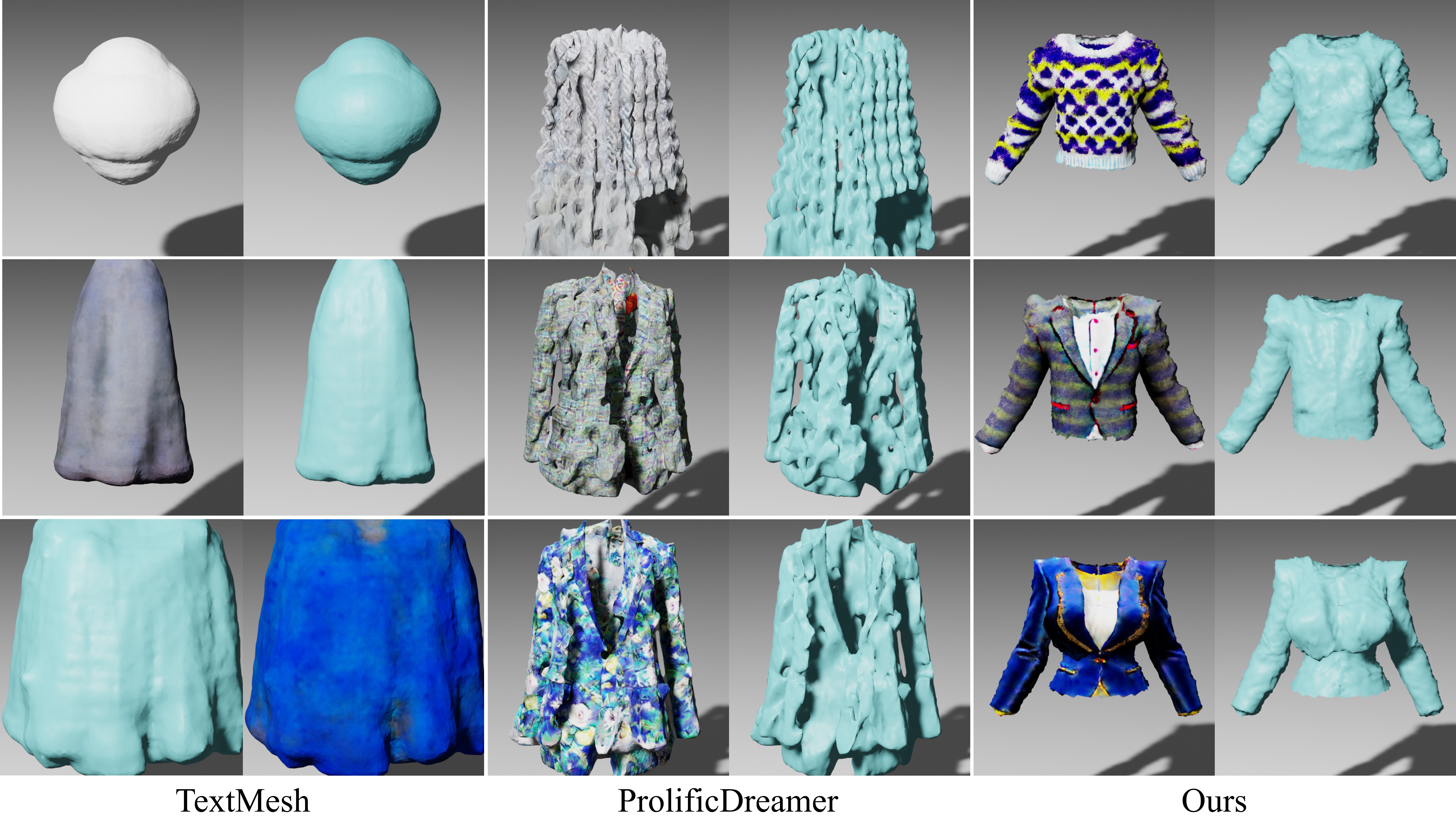}

   \caption{\textbf{Comparisons of clothes generation quality.} Garments in each row are generated using the same prompts. 
   % The clothes prompts from top to bottom are: white sweater, grey blazer, blue velvet blazer. As demonstrated from the results, meshes generated by our method significantly outperforms previous methods in texture and geometry quality.\john{Maybe don't need a floor when rendering these results.}
   }
   \label{fig:comp_clothes}
\end{figure}

\noindent \textbf{Clothes quality.}
We also evaluate the quality of the separated clothes. As demonstrated in Fig.~\ref{fig:comp_clothes}, neither TextMesh~\cite{tsalicoglou2023textmesh} or ProlificDreamer~\cite{wang2023prolificdreamer} can generate meaningful clothes shapes or textures. Besides, they cannot control the position and the size of the generated clothes, and thus cannot be fitted to an avatar. In comparison, our method can produce clothes fitting to a specific avatar and also with high-quality textures and shapes.

\begin{figure*}[t]
  \centering
   \includegraphics[width=\linewidth]{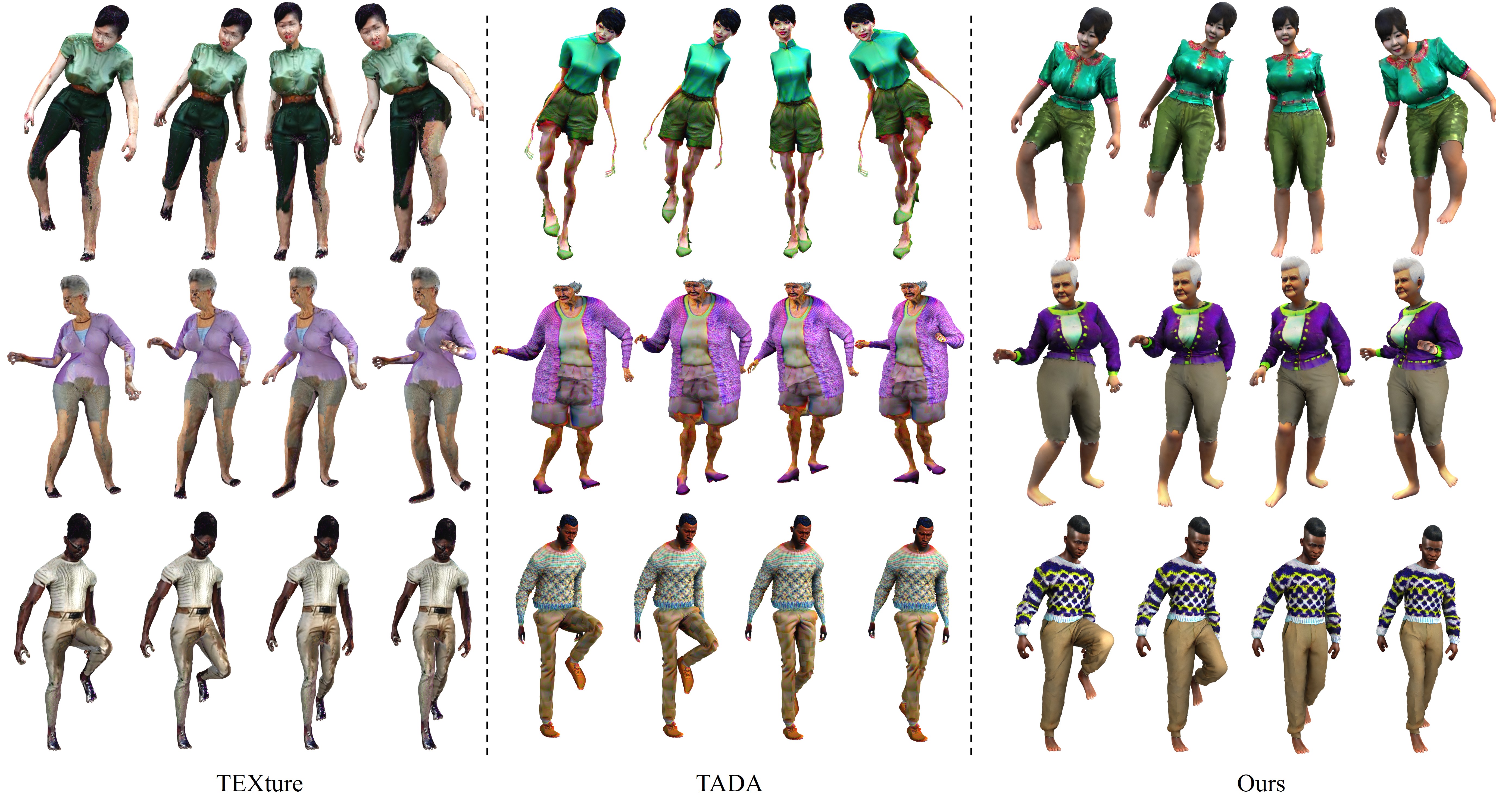}

   \caption{\textbf{Qualitative comparisons of animation results.} Our model enables the generation of disentangled human body and clothes meshes, which thus enables more photorealistic animations.}
   \label{fig:comp_motionseq}
\end{figure*}

\noindent \textbf{Animation quality.}
Our generated human body mesh is built upon the SMPL-X model and thus allows us to use SMPL-X parameters to animate the body. On the other hand, the separately generated clothes can be used in a simulation environment~\cite{MarvelousDesigner, CLO3d} to simulate their motions. We used motion sequences from AMASS~\cite{AMASS:ICCV:2019} and AIST++~\cite{li2021learn} to animate the generated avatars and simulate the clothes animations in the MarvelousDesigner~\cite{MarvelousDesigner}.  Note that the recent progress in human motion synthesis~\cite{dou2023c, cong2024laserhuman, tevet2022human, wan2023tlcontrol, zhou2023emdm,alexanderson2023listen} or human motion capture from visual signals~\cite{li2021hybrik, chen2022learning, zhang2023learning, dou2023tore, wang2023zolly,shi2020motionet} can also be employed to drive the generated character.
The animation results of TADA!, TEXTure, and our method are shown in Fig.~\ref{fig:comp_motionseq}. For TADA! and TEXTure, the clothes are entangled on the generated avatar, which leads to inconsistent wrinkles and shadows in animation. Our method disentangles clothes and human bodies, enabling realistic clothes-body interactions and visual effects.
% Since the clothes in both TADA\cite{liao2023tada} and TEXture\cite{richardson2023texture} are attached to the human body rigidly, there are no clothes-human interactions. On the other hand, our method disentangles clothes and human bodies, enabling realistic clothes-body interactions and clothes self-interactions with fine details. Besides, pose-dependent \& view-dependent visual effects such as wrinkles and shadows are baked into the learned textures of TADA\cite{liao2023tada} and TEXture\cite{richardson2023texture} and thus stay unchanged during animations, leading to unnatural visual artifacts under different poses and lighting conditions.

\begin{table}
\centering
\setlength{\tabcolsep}{2.5pt}
\begin{tabular}{c|cccc}
\toprule
Methods &  AvatarCLIP~\cite{hong2022avatarclip}  &  TEXture~\cite{richardson2023texture}  & TADA~\cite{liao2023tada} & Ours   \\ \midrule
FID $\downarrow$ &  166.0  &  132.8 &  90.2 & \textbf{84.9} \\
CLIP-FID $\downarrow$ & 34.9  & 42.8 & 27.4 & \textbf{26.2} \\
CLIP Score $\uparrow$ & 19.2 & 23.6 & 24.9 & \textbf{28.4} \\
\bottomrule
\end{tabular}
% }
\caption{Quantitative comparisons of generated ``A''-pose avatar quality in FID, CLIP-FID, and CLIP score.}
\label{table:quantitative_comps}
% \end{minipage}
% \vspace{-15pt}
\end{table}

\begin{table}%[b]{0.75\textwidth}
\centering
\setlength{\tabcolsep}{2.5pt}
\begin{tabular}{c|ccc}
\toprule
Favourite (\%, $\uparrow$)   &  TEXture~\cite{richardson2023texture}  & TADA~\cite{liao2023tada} & Ours   \\ \midrule
Texture \& Geometry Quality &  8.1  &  8.4 & \textbf{83.4} \\
Animation Realism  & 3.4  & 7.0 & \textbf{89.7} \\
Prompt Alignment  & 5.5 & 7.7 & \textbf{86.8} \\
\bottomrule
\end{tabular}
\caption{User study from 26 users in terms of visual quality, animation quality, and prompt alignment preference. The number means the percentage of users favoring a specific method. }
\label{table:user_study}
% \vspace{-15pt}
\end{table}%

% \hfill % Space between minipages, if needed
% Second minipage for the second table (duplicate)
% \begin{minipage}[b]{0.75\textwidth}

% \centering
% \resizebox{\textwidth}{!}{%

\subsection{Quantitative Comparisons}
Accurate evaluation of text-to-3D generation is a challenging task, which is rarely discussed in existing works~\cite{wu2024gpt}. Here, we use 3 metrics to evaluate generation fidelity: FID~\cite{heusel2017gans, parmar2021cleanfid}, CLIP-FID~\cite{kynkaanniemi2022role} and CLIP Score~\cite{hessel2021clipscore}. CLIP Score is the CLIP feature distance between the input prompts and the rendered images of the generated avatars. For FID and CLIP-FID, we measure the distance between the rendered images from our generated 3D avatars, and images generated from Stable Diffusion~\cite{rombach2021highresolution} using the same prompts. We evaluate all methods using 30 prompts and render 300 views for each prompt, resulting in 9000 images. As shown in Table~\ref{table:quantitative_comps}, our method achieves the best results in all 3 metrics compared with previous methods, demonstrating the high visual quality and textual alignment of our pipeline.

\subsection{User Study}
\label{sec:user_study}
We also conduct a comprehensive user study to evaluate both the generation quality and the animation capability of our pipeline. We design a questionnaire and present 16 full-body animations made by TEXture~\cite{richardson2023texture}, TADA!~\cite{liao2023tada}, and our method. 
% To generate these animations, we randomly picked 16 different prompts for clothed humans. 
The survey is conducted among 26 researchers and artists from either academia or industry. The participants are asked to pick the one with the best semantic consistency, visual quality, and animation quality. Note that we ask participants to exclude animation quality when evaluating texture\& geometry quality. As shown in Table ~\ref{table:user_study}, our model is the most preferred one in all aspects of generation quality. 

% \begin{table}[t]
% %\vspace{-1.0 em}
% \centering
% \setlength{\tabcolsep}{2.5pt}
% \resizebox{\linewidth}{!}{
% \begin{tabular}{c|ccc}
% \toprule
% Favourite (\%, $\uparrow$)   &  TEXture~\cite{richardson2023texture}  & TADA~\cite{liao2023tada} & Ours   \\ \midrule
% Texture \& Geometry Quality &  8.1  &  8.4 & \textbf{83.4} \\
% Animation Realism  & 3.4  & 7.0 & \textbf{89.7} \\
% Prompt Alignment  & 5.5 & 7.7 & \textbf{86.8} \\
% \bottomrule
% \end{tabular}
% }

% \caption{\textbf{User study} from 26 users in terms of visual quality, animation quality, and prompt alignment preference. The number means the percentage of users selecting a specific method. 
% % User feedbacks show that our generation pipeline markedly surpasses other baseline models in aspects geometry and texture, animation quality, and alignment with the provided input prompt.
% }
% %\vspace{-1.5 em}
% \label{table:user_study}
% \end{table}

\subsection{Ablation Study}
\label{sec:ablation}
\begin{figure}[htbp]
    \centering
    \begin{minipage}{0.48\textwidth}
        \includegraphics[width=\linewidth]{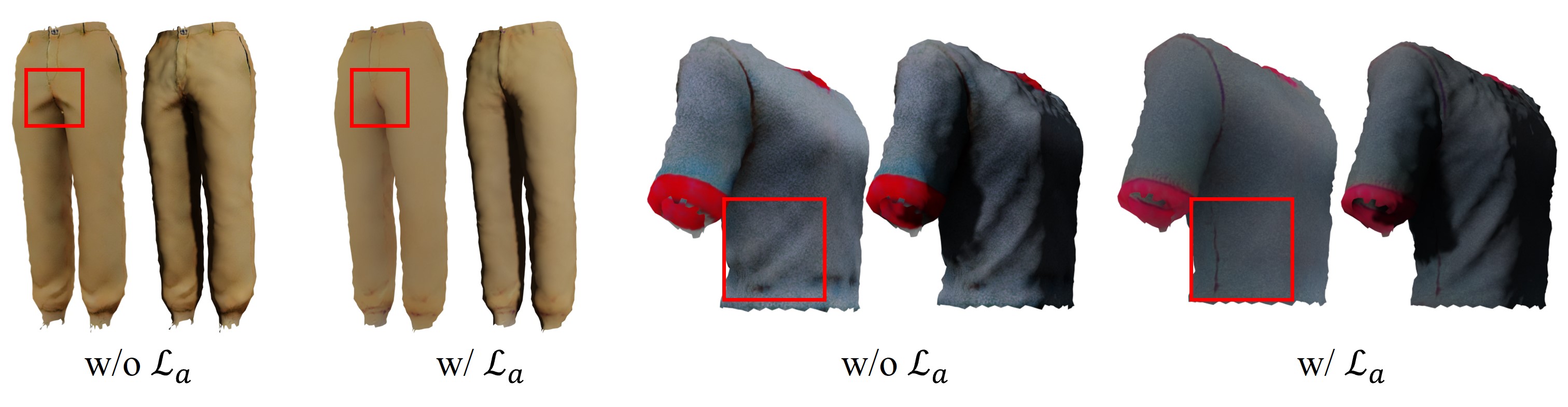}
    \end{minipage}
    \hfill
    \begin{minipage}{0.48\textwidth}
        \includegraphics[width=\linewidth]{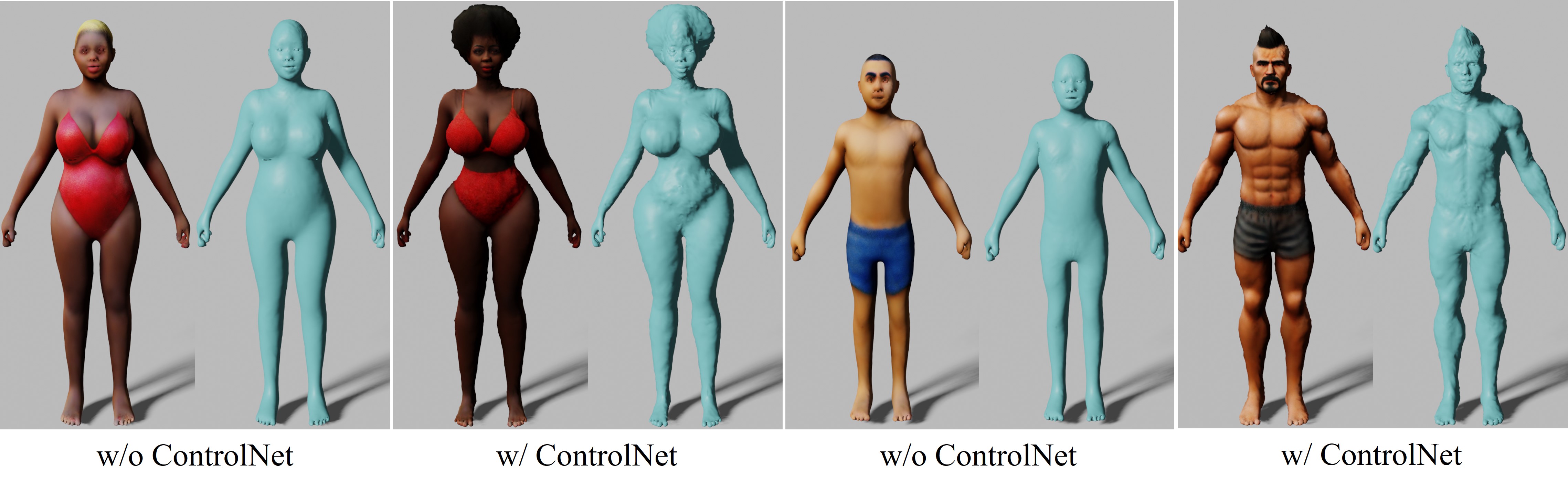}
    \end{minipage}
    \caption{Ablation study on our shading model (left) and ControlNet as guidance (right).}
    \label{fig:ablation}
\end{figure}

We first evaluate the effects of our albedo smoothness constraints and 2D-pose ControlNet guidance. In Fig.~\ref{fig:ablation}, we compare the rendered albedo and RGB images w/ or w/o our albedo smoothness constraint $\mathcal{L}_a$. As can be seen from the visual results, the albedo generated without $\mathcal{L}_a$ shows baked-in shadows and wrinkles, which leads to unsatisfactory visual artifacts in animations. In comparison, using $\mathcal{L}_a$ largely alleviates the problem in the albedo while still enabling the correct rendering of the shadows and wrinkles on RGB images.
% Such baked in textures could result in unrealistic visuals artifacts under different lighting conditions and during animations. 
% \begin{figure}[htbp]
%   \centering
%    \includegraphics[width=\linewidth]{assets/sec5/ablation_albedo_1118.jpg}

%    \caption{\textbf{Ablation study on our shading model.} For each garment, we showcase the albedo map (left) and the rendered RGB image (right). Our full shading model with albedo smoothness constraint $\mathcal{L}_a$ enables the rendering of shadows in wrinkles while alleviating baked-in artifacts of shadows in the albedo.}
%    \label{fig:ablation_albedo}
% \end{figure}

In Fig.~\ref{fig:ablation}, we validate the effectiveness of OpenPose-based ControlNet as guidance during our human body generation. As can be observed from the results, generated avatars without ControlNet have over-smoothed textures and geometry. In contrast, our utilization of ControlNet produces significantly finer details in both geometry and texture. 
% More ablation studies on simultaneously distilling clothes and clothed avatars are included in the supplementary material.

\begin{figure}[htbp]
  \centering
   \includegraphics[width=\linewidth]{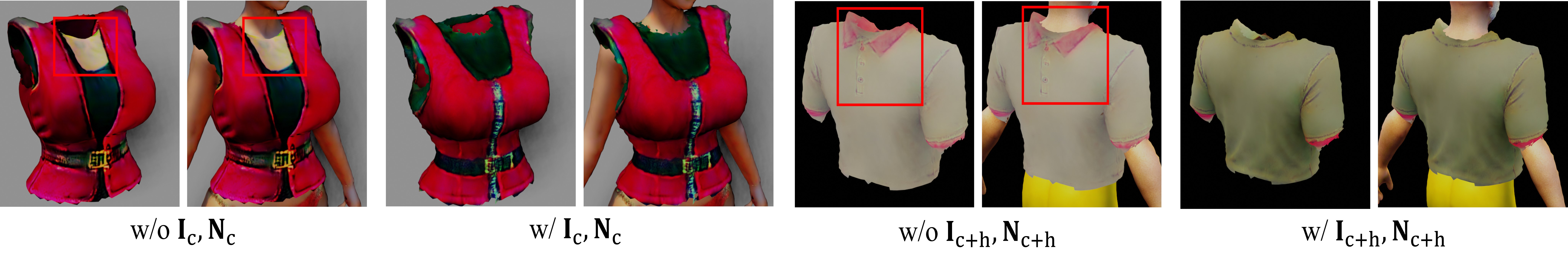}

   \caption{Ablation study on guidance from clothes renders$\{ \mathbf{I}_{\text{c}}, \mathbf{N}_{\text{c}} \}$ and clothed human renders $\{\mathbf{I}_{\text{c+h}}, \mathbf{N}_{\text{c+h}}\}$.}
   \label{fig:ablation_icnc}
   % \vspace{-15pt}
\end{figure}

In Fig.~\ref{fig:ablation_icnc} we demonstrate the necessity of using both guidances from clothes renders, $\{\mathbf{I}_{\text{c}}, \mathbf{N}_{\text{c}}\}$, and clothed human renders, $\{ \mathbf{I}_{\text{c+h}}, \mathbf{N}_{\text{c+h}}\}$. Applying $\{ \mathbf{I}_{\text{c+h}}, \mathbf{N}_{\text{c+h}}\}$ without clothes leads to the incorrect generation of skin colors on the clothes meshes, as seen on the left of Fig.~\ref{fig:ablation_icnc}. On the other hand, with only clothes guidance, the garments are prone to learn texture and semantics incoherent with the human body, such as the back collar on the right of Fig.~\ref{fig:ablation_icnc}.

% \begin{figure}[htbp]
%   \centering
%    \includegraphics[width=\linewidth]{assets/sec5/ablation_controlnet_lr.jpg}

%    \caption{\textbf{Ablation study on ControlNet as guidance.} 
%    % We compared the generated avatars with the Stable Diffusion model and ControlNet as guidance, respectively. 
%    We compare the avatars generated by the ControlNet with the ones generated by the vanilla Stable diffusion model.
%    The results showed that the ControlNet model with 2D pose conditions generates avatars with more detailed geometry and texture.}
%    \label{fig:ablation_ctrl}
% \end{figure}

% \begin{figure}[htbp]
%   \centering
%   % First figure in the first minipage
%   \begin{minipage}{.48\linewidth}
%     \centering
%     \includegraphics[width=\linewidth]{assets/sec5/ablation_controlnet_lr.jpg}
%     \caption{Ablation study on ControlNet as guidance.
%     %We compare the avatars generated by the ControlNet with the ones generated by the vanilla Stable diffusion model. The results showed that the ControlNet model with 2D pose conditions generates avatars with more detailed geometry and texture.
%     }
%     \label{fig:ablation_ctrl}
%   \end{minipage}\hfill
%   % Second figure in the second minipage
%   \begin{minipage}{.48\linewidth}
%     \centering
%     \includegraphics[width=\linewidth]{assets/sec5/ablation_albedo_1118.jpg}
%     \caption{Ablation study on our shading model.
%     %For each garment, we showcase the albedo map (left) and the rendered RGB image (right). Our full shading model with albedo smoothness constraint $\mathcal{L}_a$ enables the rendering of shadows in wrinkles while alleviating baked-in artifacts of shadows in the albedo.
%     }
%     \label{fig:ablation_albedo}
%   \end{minipage}
% \end{figure}

% fig.1 use learnable mask / not use learnable mask

% \begin{figure}[t]
%   \centering
%    \includegraphics[width=\linewidth]{assets/sec5/ablation_mask_lr.jpg}

%    \caption{\textcolor{red}{Still need tuning...} Diverse shapes generated from our learnable mask. As could be seen from the results, our method is able to generate clothes with different details like collar shapes and strap width.}
%    \label{fig:ablation_mask}
% \end{figure}

\subsection{Applications}
\label{sec:application}
\begin{figure}[htbp]
    \centering
    \begin{minipage}{0.75\textwidth}
        \includegraphics[width=\linewidth]{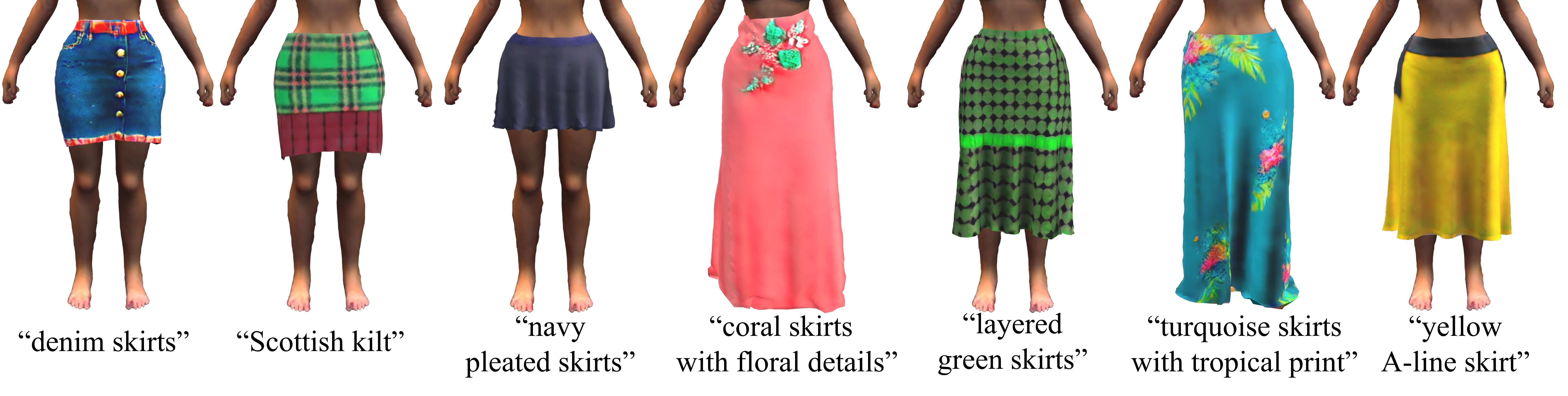}
    \end{minipage}
    \hfill
    \begin{minipage}{0.75\textwidth}
        \includegraphics[width=\linewidth]{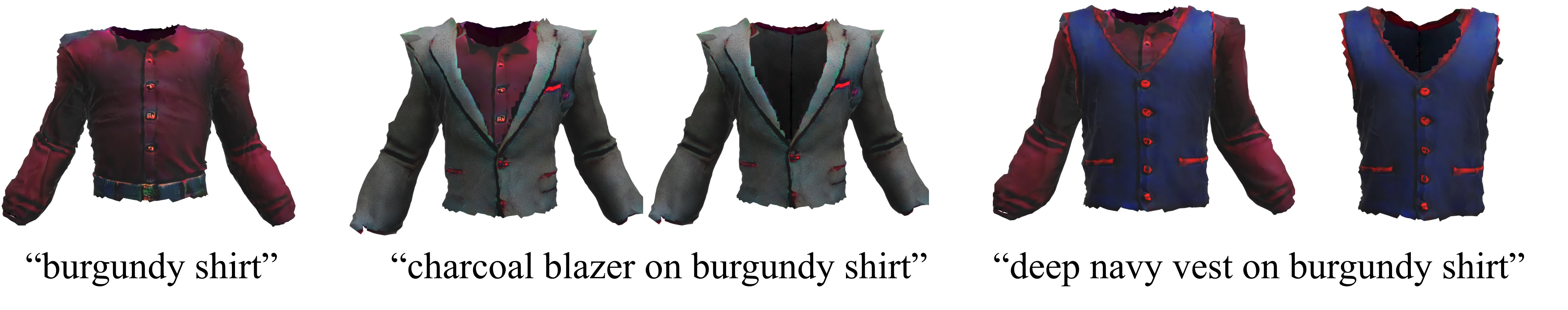}
    \end{minipage}
    \caption{\textbf{Complex garments generation.} Our pipeline could be generalized to generate more complex garments like skirts and multi-layered garments.}
    \label{fig:complex}
\end{figure}

\noindent \textbf{Complex garments generation}.
Our SO-SMPL design can also be generalized to more complex garment types. For instance, by adding additional layers of offsets on top of the first clothing layer, we can generate more complex multi-layer garments, as shown in Fig.~\ref{fig:complex}. 
Besides, leveraging BCNet~\cite{jiang2020bcnet}, a parametric skirts model conditioned on SMPL parameters, our pipeline can be extended to generating clothes that do not follow SMPL-X's topological structure such as skirts and dresses, as depicted in Fig.~\ref{fig:complex}. More details can be found in the supplementary material.

\begin{figure}[t]
  \centering
   \includegraphics[width=0.8\linewidth]{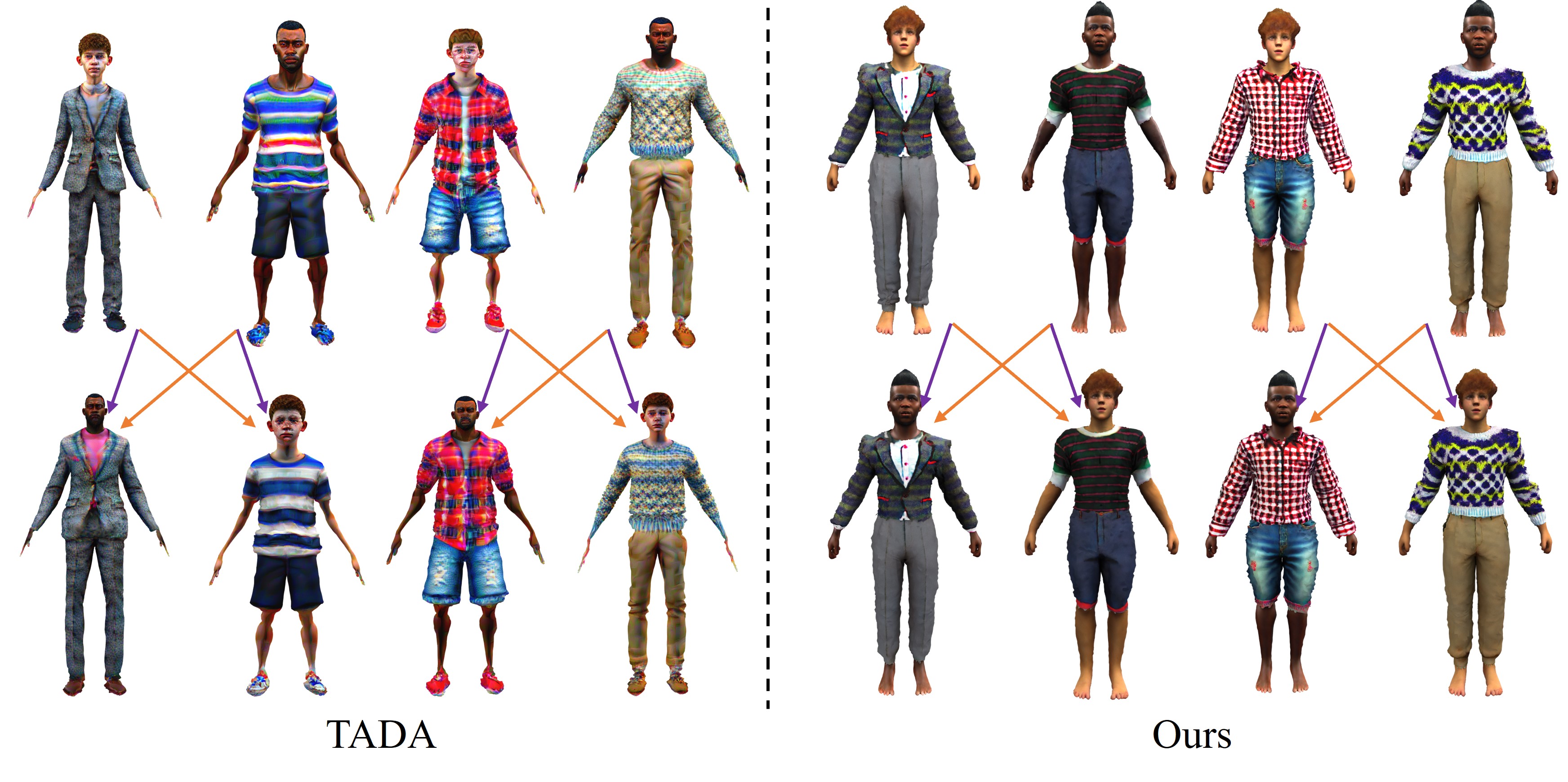}

   \caption{\textbf{Comparisons in virtual try-on applications.}
   The orange and purple arrows represent compositions of the human body and the clothes, respectively. Changing clothes in TADA!~\cite{liao2023tada} often leads to undesirable changes in human identity or clothes while changing clothes in our method is more photorealistic and keeps the same human identity and clothes.
   }
   \label{fig:comp_tryon}
\end{figure}

\noindent \textbf{Virtual try-on.} Our disentangled human-clothes representation inherently allows us to change the outfit of a certain avatar, or put the same clothes on different avatars. 
As depicted in Fig.~\ref{fig:comp_tryon}, while previous methods like TADA!~\cite{liao2023tada} also enable virtual try-on, they struggle to maintain consistency in clothing and human identity during the try-on process. As can be seen from the left half of Fig.~\ref{fig:comp_tryon}, when changing clothes for the white teenage boy with TADA!, the facial details and body shape are changed; on the other hand, the colors and patterns of the clothes also changed when putting the same clothes on different avatars with TADA!. On the contrary, our disentangled representation supports fixing the human identity and clothes details when switching outfits.

\section{Limitations\&Conlusion}
\noindent \textbf{Limitations.}
Despite promising results, our pipeline still has limitations. 
Since images generated by diffusion models are of final shadings, it is difficult to get fully shadow-free material decomposition even with our shading model and albedo smoothness constraint. Besides, our results still suffer from cartoonish colors and over-smoothing textures, which is a common problem for distillation-based frameworks.
% Currently, since our SO-SMPL representation is based on SMPL-X model, our pipeline could not generate clothes types that do not follow SMPL-X's topological structure, such as skirts and dresses. To extend to such clothes, we may follow \cite{jiang2020bcnet} and utilize a parametric dress/skirt model as a template instead of the original SMPL-X. % should think about a limitation
% Furthermore, since the clothes region mask $\mathbf{M}_{\text{c}}$ is pre-defined and not learnable, our method could not generate diverse clothes patterns such as thin straps or asymmetric collars. 
Another limitation is the over-saturation of colors in our distillation-based framework. Although distillation techniques\cite{wang2023prolificdreamer} have aimed to address this issue, it remains a challenge. Our offset-based representation also can not handle avatars with long hair, which requires advanced 3D representations\cite{sklyarova2023haar}.
What's more, our generated clothes do not follow any sewing patterns and are without any physical attributes such as stretch, bending, and friction. Therefore, generating garments with sewing patterns is a promising direction for exploration. We leave these potential tasks to future works. 

%\section{Conclusion}
\noindent \textbf{Conclusion.}
In this paper, we have presented an innovative approach for generating human avatars through our Sequentially Offset-SMPL (SO-SMPL) representation from textual descriptions. To the best of our knowledge, our pipeline is the first to produce avatars in a disentangled manner: first generate a human body mesh then a clothes mesh on top of it. The 3D avatars and clothes generated by our pipeline exhibit remarkable diversity and high fidelity in both texture and geometric detailing. They could also be easily utilized in CG software for animation and simulation, thereby opening up exciting possibilities in fields such as virtual reality, gaming, and digital fashion. Experiments demonstrate that our pipeline outperforms existing text-to-3D generation methods in texture\&geometry quality, alignment with text descriptions, and animation quality.

\clearpage  % TODO FINAL: This \clearpage needs to be removed from both review and camera-ready versions.

\section*{Acknowledgements}
We thank Tingting Liao, and Yukang Cao for their insightful suggestions. We also appreciate Ziyu Chen for fruitful discussions during the research process.

\noindent This work was partly supported by the Fundamental Research Funds for the Central Universities, 111 Project, China under Grants B07022 and (Sheitc) 150633, and Shanghai Key Laboratory of Digital Media Processing and Transmissions, China, and U.S. National Science Foundation CBET-2115405. This work is also partly supported by the Innovation and Technology Commission of the HKSAR Government under the ITSP-Platform grant (Ref:  ITS/319/21FP) and the InnoHK initiative (TransGP project).

% is also supported by the Innovation and Technology Commission of the HKSAR Government under the InnoHK initiative.

% ---- Bibliography ----
%
% BibTeX users should specify bibliography style 'splncs04'.
% References will then be sorted and formatted in the correct style.
%
\bibliographystyle{splncs04}
\bibliography{main}

\clearpage
\setcounter{page}{1}

\newpage
\begin{center}
    \Large
    \textbf{Disentangled Clothed Avatar Generation from Text Descriptions}\\[0.5em] % Replace with your title
    Supplementary Material\\[1.0em]
\end{center}

% TODO in supp
% 1. Edit the human body's betas parameters [x]
% 2. implementation details [x]
% 3. generalize to shoes(mention it) a little [x]
% 4. mesh subdivision [x]
% 5. about the user survey, and details about the questions. also, FID details. [x]
% 6. Prompt details, what kind of extra prompts are used [x]
% 7. dress details  [x]
% 8. Negative impact [x]
% 9. same human, different clothes(in video) []
% 10. CVPR 2024 comps [] % what to do with humannorm
 
\section{Pipeline Details}
\subsection{Garment Types}
\begin{figure}[htbp]
  \centering
   \includegraphics[width=0.8\linewidth]{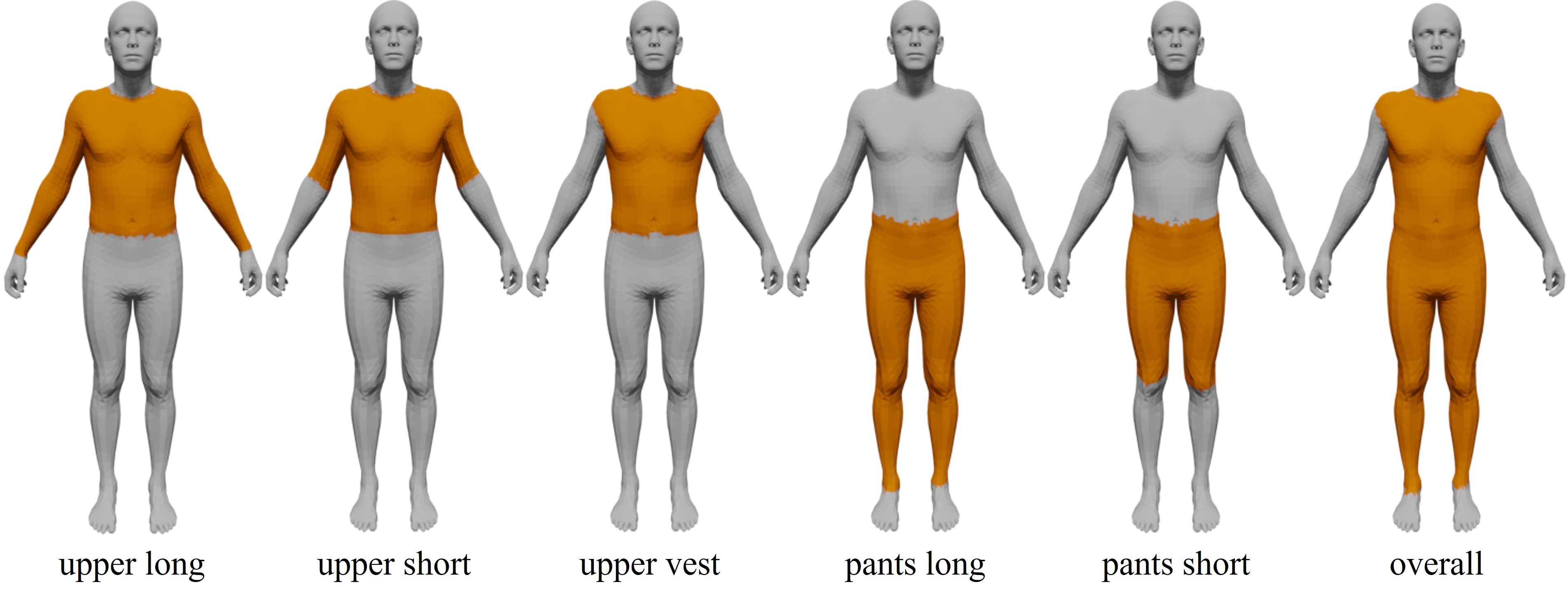}

   \caption{\textbf{Mask templates.}}
   \label{fig:supp_mask}
\end{figure}

As mentioned in Sec.~\ref{sec:geometry}, we utilized SMPL-X\cite{SMPL2015} part segments to craft templates for 6 different types of clothes. The types of masks can be seen in Fig.~\ref{fig:supp_mask}. Note that our pipeline could be generalized to other shapes of garments by defining new masks, e.g. we could define a mask for feet areas to generate shoes.

As mentioned in the main paper, we could also leverage templates from BCNet~\cite{jiang2020bcnet} to generate skirts and short skirts. Specifically, we used the provided base template mesh along with its PCA-based linear displacement deformation. The PCA coefficients for the deformation could be produced through a two-layer MLP network conditioned on human body shape parameter $\beta$. As a result, a fitting skirt mesh template is acquired and could be used in our pipeline for offsets and texture optimization. 

% \subsection{Offset Settings}
% The details of offset settings, and how to ensure the offset is bigger than 0 (seems too detailed).
\subsection{Prompting}
% The distinguishment between clothing and the human body is ambiguous in the descriptions by text prompts. 
Besides the input prompts, we also added extra descriptions after the prompts to enhance the performance. For human body generation, we added \textit{photorealistic, ultra-detailed, 8k uhd}. For garments, we added \textit{wrinkleless smooth and flat}. These descriptions are chosen empirically. 

Simply using descriptive prompts can not guarantee the distinction between the human body and clothes. For instance, even with descriptions like ``unclothed'' or ``without accessories'', there could still be undesired extras on the human body. On the other hand for garment generation, the generated clothes sometimes contain undesired ``human parts'' like skin or even limbs. To address this problem, we added negative prompts for both bodies (e.g. ``loose clothes, accessories'') and clothes generation(e.g. ``skin, legs''), to encourage the predicted noise from Stable Diffusion to guide the model away from such entanglement. Specifically, the original classifier-free guidance\cite{ho2022classifier} against unconditioned noise $\epsilon_{\phi}(\mathbf{I}_t; t)$ is:
\begin{equation}
\hat{\epsilon}_{\phi}(\mathbf{I}_t; \mathbf{y}, t) = (1+\omega)\epsilon_{\phi}(\mathbf{I}_t; \mathbf{y}, t) - \omega\epsilon_{\phi}(\mathbf{I}_t; t)
\end{equation}
and the modified version with negative prompting is:
\begin{equation}
\hat{\epsilon}_{\phi}(\mathbf{I}_t; \mathbf{y}, t) = (1+\omega)\epsilon_{\phi}(\mathbf{I}_t; \mathbf{y}, t) - \omega\epsilon_{\phi}(\mathbf{I}_t; \mathbf{y}_{\text{n}}, t)
\end{equation}

\subsection{Mask Blending}
To render the clothed human rgb image $\mathbf{I}_{\text{c+h}}$ from separate body and clothes textures, we rasterize the SO-SMPL with its vertex-level mask $\mathbf{M}_{\text{c}}$ to obtain 2D blending mask $\mathbf{I}^{\text{mask}}$ and use it to combine the rgb values of human body $\mathbf{I}_{\text{h}}$ \& clothing $\mathbf{I}_{\text{c}}$, i.e. :
\begin{align}
\begin{split}
 \mathbf{I}_{\text{c+h}}   &= \mathbf{I}_{\text{c}} \odot \mathbf{I}^{\text{mask}} + \mathbf{I}_{\text{h}} \odot (\mathbf{1} - \mathbf{I}^{\text{mask}})
\end{split}
\label{equ:supp_blend}
\end{align}
where $\odot$ denotes the Hardmard product. Notably, self-occlusions are already considered in the rasterization process thus $\mathbf{I}^{\text{mask}}$ could be applied directly. 

% \subsection{Camera settings}
% \subsubsection{Semantic Zoom-ins}
% \subsubsection{Adaptive Camera}
% During the avatar generation, the human body shape parameter $\beta$ could change fluctuate drastically.

\section{Additional Experiments}
\begin{figure*}[htbp]
  \centering
   \includegraphics[width=\linewidth]{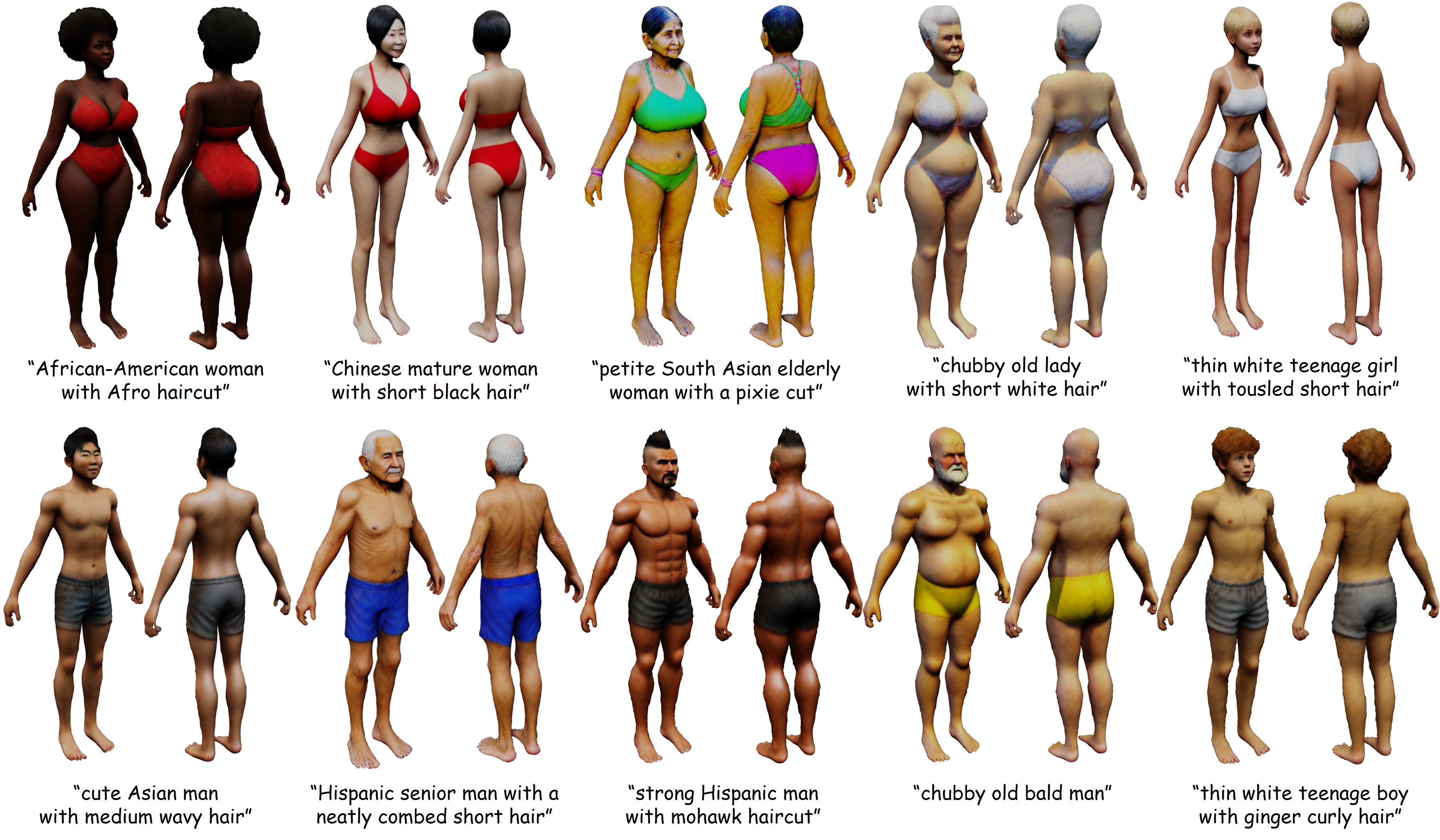}

   \caption{\textbf{Generated human body meshes.}}
   \label{fig:supp_body}
\end{figure*}

\begin{figure*}[t]
  \centering
   \includegraphics[width=\linewidth]{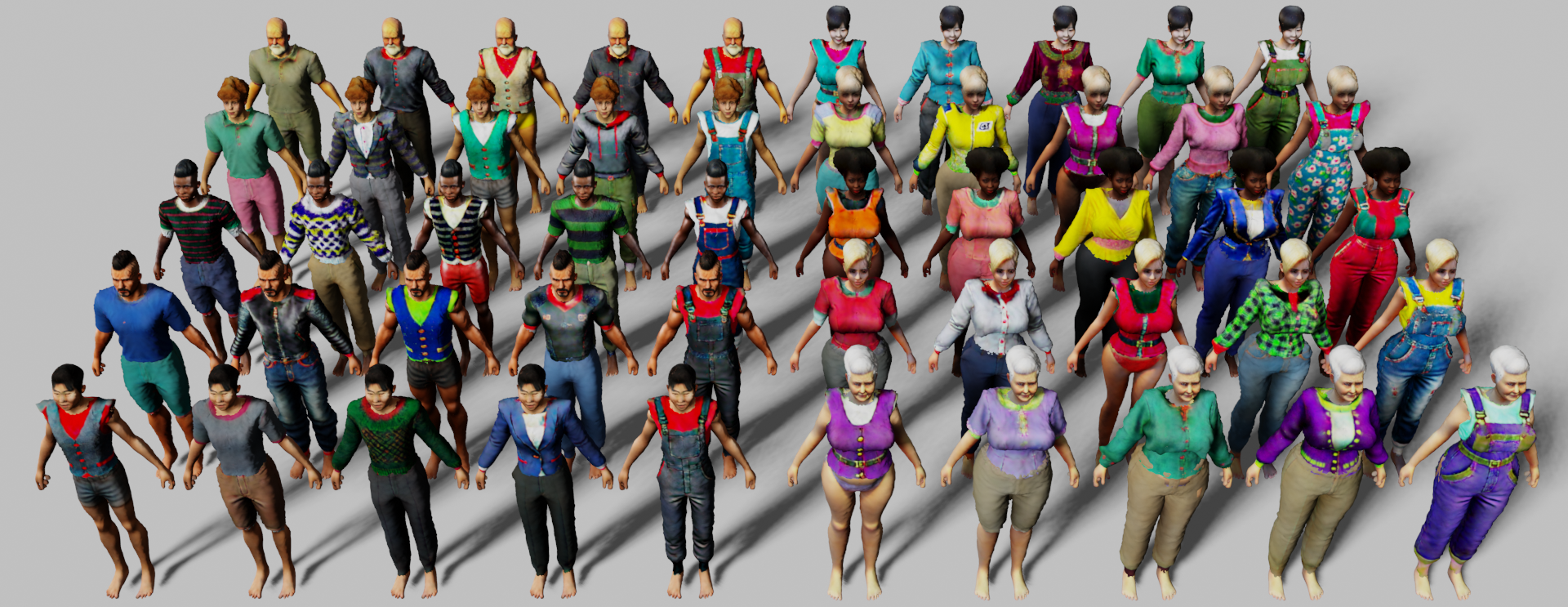}

   \caption{\textbf{An overall view of our generated clothes and avatars.}}
   \label{fig:supp_all}
\end{figure*}

\begin{figure*}[t]
  \centering
   \includegraphics[width=\linewidth]{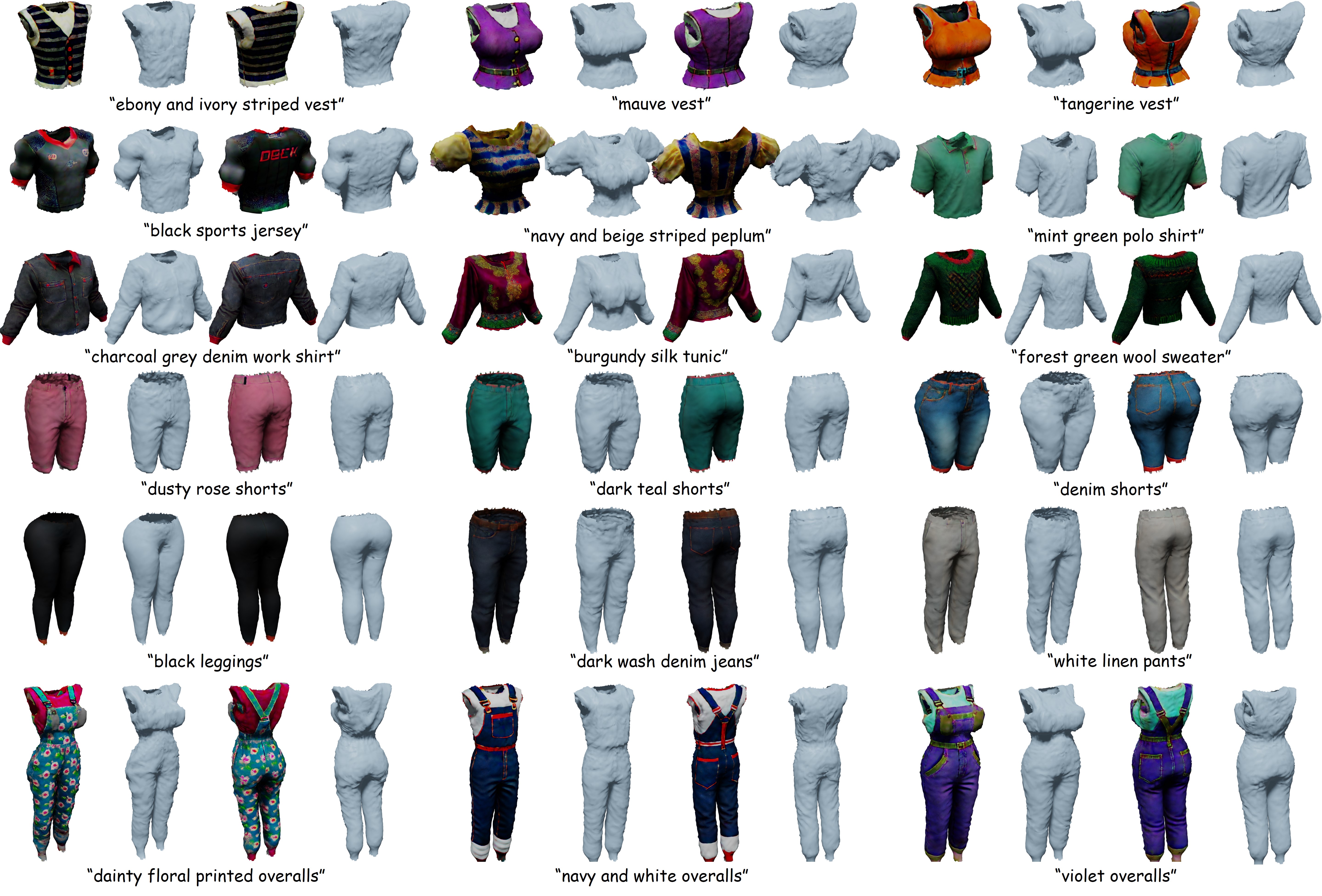}

   \caption{\textbf{Generated clothes gallery.}}
   \label{fig:supp_gallery}
\end{figure*}

\subsection{Implementation Details}
All experiments for this paper are conducted on a single GeForce RTX 4090 GPU. For each human body mesh, we optimize our model for 15000 steps, which takes around 12GB of GPU memory and 2.5 hours to generate. As for clothes, we optimize them for 12000 steps, which takes around 19GB of GPU memory and 2.5 hours of optimization. For human body optimization, we randomly sample around the center of the human body with an azimuth angle of [-180$^\circ$, 180$^\circ$] within a distance range of [1.25, 2.3], with a camera FoV of [45$^\circ$, 50$^\circ$]. We also add semantic close-ups on the face and hands region to enhance the fidelity of details. We adopt a similar strategy for clothes optimization, except that we introduce a position bias to coup with different garment types, e.g. move up the camera for upper garments. We set a learning rate of 0.0003 for offsets, 0.005 for texture, and 0.003 for body shape $\beta$. We also subdivide the original SMPL-X mesh to increase our model's ability to represent geometric details, following TADA~\cite{liao2023tada}.

\subsection{Evaluation Details}
\begin{figure*}[htbp]
  \centering
   \includegraphics[width=\linewidth]{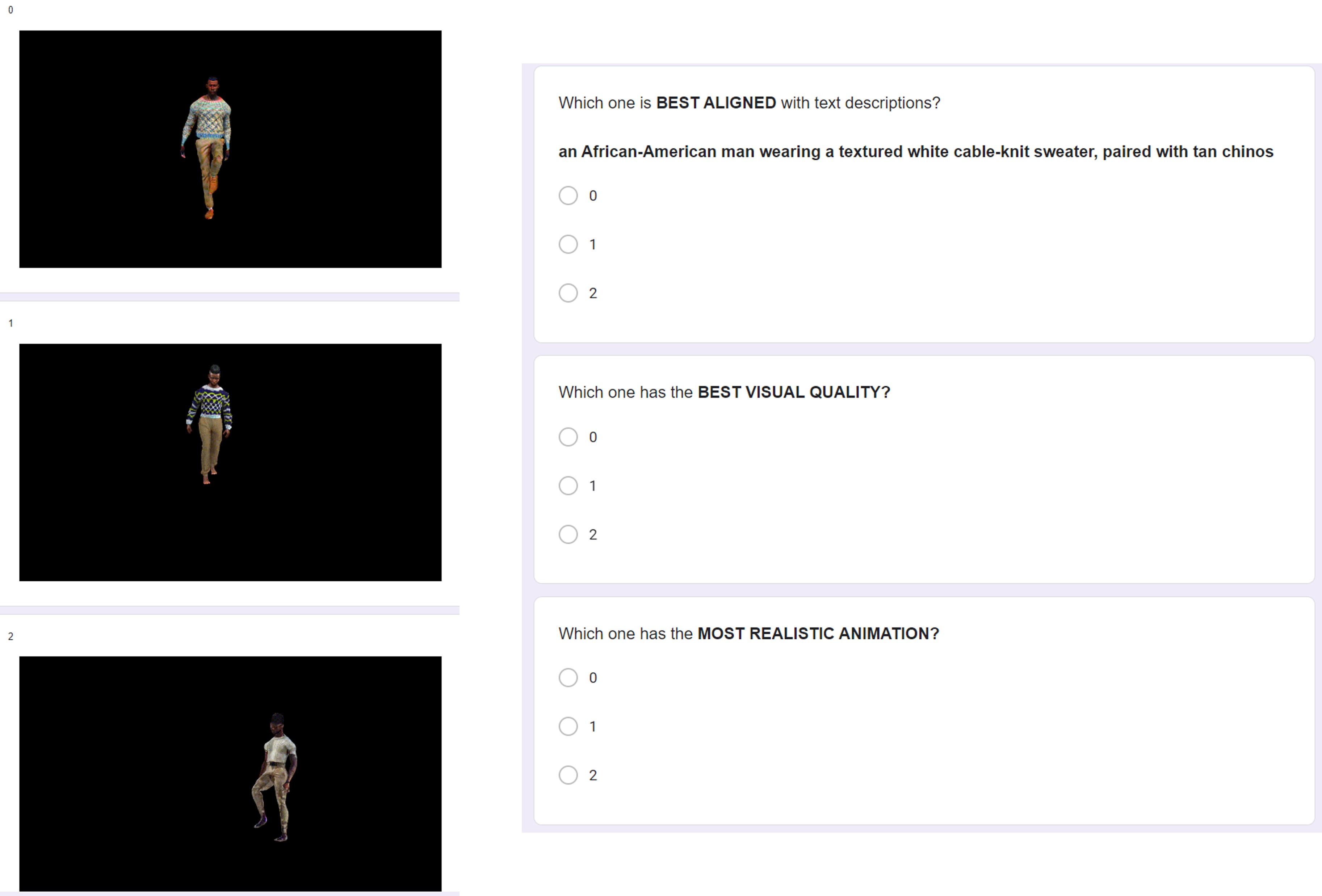}

   \caption{Demonstration of our user study form.}
   \label{fig:supp_survey}
\end{figure*}

We use a digital form to conduct our user study. We show part of our user study form in Fig.~\ref{fig:supp_survey}. Three randomly ordered animation gifs made by different methods are played simultaneously. For each animation, three questions regarding text alignment, visual quality, and animation realism were asked. The criteria for the three metrics are described at the beginning of the form, in which we explicitly instruct participants to assess the visual quality of the avatars independently of the animation quality. Besides, we also conducted an additional user study among 18 users that compares our method with AvatarCLIP~\cite{hong2022avatarclip}, in which our method is favored among all cases and categories. The avatars generated by AvatarCLIP exhibit noisy geometry and low-fidelity textures, as shown in Fig.~\ref{fig:comp_static}.

For quantitative computations, we picked 30 prompts for generation, and we randomly sampled 300 camera views centered around the generated human body center. Then, we render all generated avatars with these same views, resulting in 9000 rendered images for each method. Close-up views on faces and hands are also sampled and paired with respective prompts, i.e. \textit{the face of ...}, \textit{the hand of ...}. We then used Stable Diffusion v1.5~\cite{rombach2021highresolution} to generate 300 images for each prompt as the reference set. FID and Clean-FID between rendered images and the reference set are computed using the public tool provided by \cite{parmar2021cleanfid}. CLIP scores are measured directly between the input prompts and the rendered images using torchmetrics~\cite{torchmetrics}. In tab.\ref{table:supp_quantitative_comps}, we also added additional prompts to make a more sufficient comparison between the most competitive concurrent work with our results.

\subsection{Shape Editing}
\begin{figure}[htbp]
  \centering
   \includegraphics[width=0.8\linewidth]{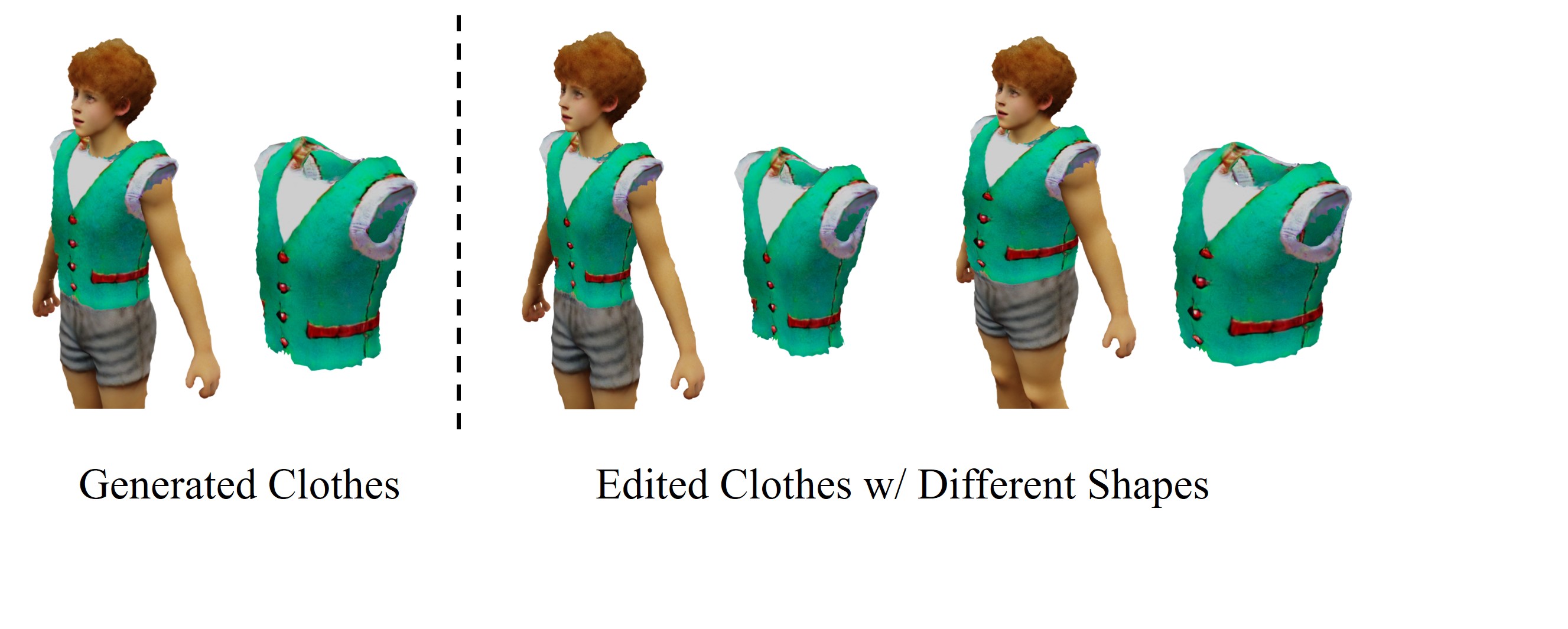}

   \caption{\textbf{Shape editing on clothes}. SO-SMPL representation allows us to edit human shapes and fit clothes to them while maintaining the original patterns and details.}
   \label{fig:shape_edit}
\end{figure}

Our SMPL+offset-based model also allows us to edit the shape of generated characters and garments. To achieve this, we could simply modify the SMPL-X shape parameter $\beta$ of a character, and then add the garment offsets $\mathbf{O}_{\text{c}}$ on top of the edited character. As shown in Fig.~\ref{fig:shape_edit}, clothes generated by our methods could be changed according to different human shapes and sizes without re-generating. This flexibility helps us to manipulate the generated human body and allows generated clothes to be re-used on different avatars while maintaining fit with their body types. 

\subsection{Avatar Gallery}
A gallery of our generated avatars is presented in Fig.~\ref{fig:supp_body}. Our method can generate highly detailed human body meshes.

\subsection{Clothes Gallery}
A gallery of our generated clothes of different types is presented in Fig.~\ref{fig:supp_gallery}. As could be seen from the results, our pipeline could generate a diverse range of clothes of different colors, materials, and types.

\subsection{Additional Quantitative Comparisons}
We added extra quantitative comparisons of ``A''-pose avatar quality on our method and the most competitive concurent work TADA~\cite{liao2023tada}. The experiment is conducted on 50 prompts with 300 renders each. As could be seen from Tab. \ref{table:supp_quantitative_comps}, our method outperforms TADA~\cite{liao2023tada} in all three metrics.
\begin{table}
\centering
\setlength{\tabcolsep}{2.5pt}
\begin{tabular}{c|cc}
\toprule
Methods &  TADA~\cite{liao2023tada} & Ours   \\ \midrule
FID $\downarrow$ & 104.1 & \textbf{98.8} \\
CLIP-FID $\downarrow$ & 28.7 & \textbf{26.2} \\
CLIP Score $\uparrow$ & 24.7 & \textbf{27.8} \\
\bottomrule
\end{tabular}
% }
\caption{Additional comparisons of generated ``A''-pose avatar quality in FID, CLIP-FID, and CLIP score.}
\label{table:supp_quantitative_comps}
% \end{minipage}
% \vspace{-15pt}
\end{table}

\subsection{Additional Qualitative Comparisons}
% Due to the nature of this field, numerous text-to-avatar pipelines are proposed.\john{Should we compare with DreamHuman? It is not open source but has an animation gallery.} 

\begin{figure*}[htbp]
  \centering
  \includegraphics[width=\linewidth]{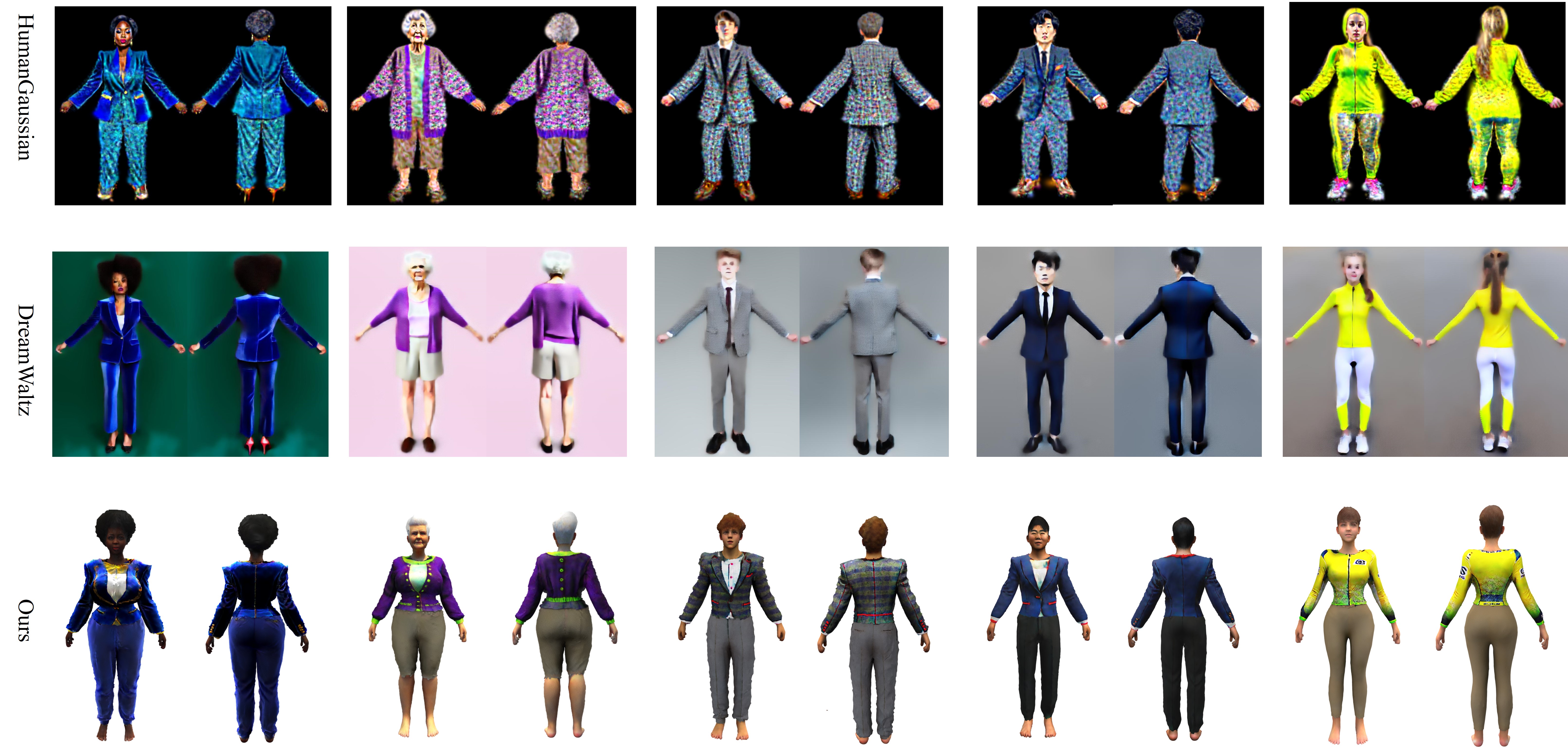}

   \caption{\textbf{Comparisons of ``A-pose'' quality with HumanGaussian~\cite{liu2023humangaussian} and DreamWaltz~\cite{huang2023dreamwaltz}.} }
   \label{fig:supp_comp_static_dw}
\end{figure*}

\begin{figure*}[htbp]
  \centering
  \includegraphics[width=\linewidth]{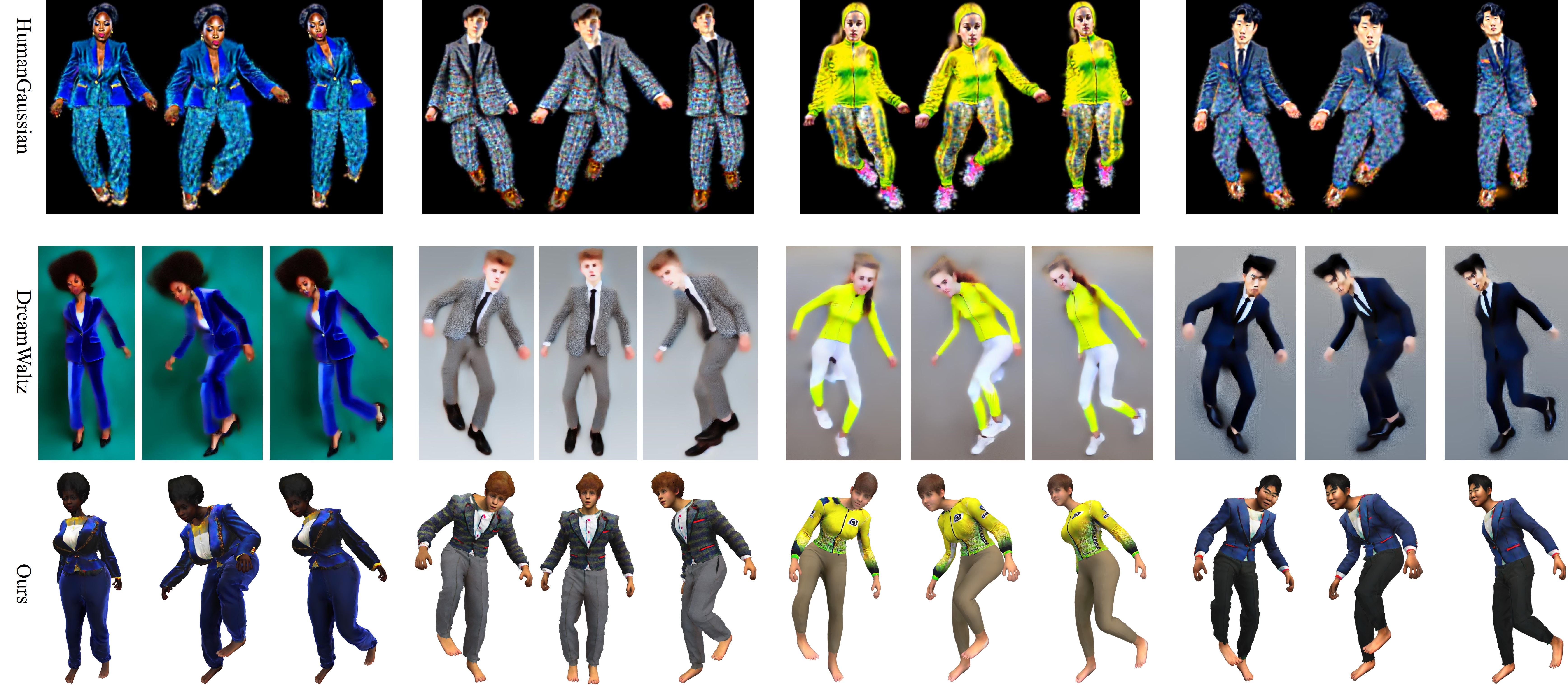}

   \caption{\textbf{Comparisons of animation quality with HumanGaussian~\cite{liu2023humangaussian} and DreamWaltz~\cite{huang2023dreamwaltz}.} For a better visual comparison, please refer to our supplementary videos.}
   \label{fig:supp_comp_motion_dw}
\end{figure*}

We present comparisons with additional state-of-the-art animatable text-to-avatar methods, namely DreamWaltz~\cite{huang2023dreamwaltz} and HumanGaussian~\cite{liu2023humangaussian}. Visual comparisons are presented in Fig.~\ref{fig:supp_comp_motion_dw} and Fig.~\ref{fig:supp_comp_static_dw}. 

As seen in Fig.~\ref{fig:supp_comp_static_dw}, DreamWaltz~\cite{huang2023dreamwaltz} is capable of generating photorealistic plausible results, yet it suffers from over-smooth texture and noises due to its NeRF-based representation. Consequently in Fig.~\ref{fig:supp_comp_motion_dw}, noises move along with the joint motions, causing severe visual artifacts. A more recent work HumanGaussian~\cite{liu2023humangaussian} also suffers from geometric artifacts for its point-based Gaussian representation, resulting in noisy textures and motions.

\section{Negative Impact}
The advancement of text-driven avatar and clothing generation technologies carries with it potential risks such as privacy violations, intellectual property infringement, and the propagation of misinformation. Our human body generation method is particularly susceptible to malicious applications, which raises significant ethical concerns. We emphasize the necessity for transparency in further development and applications of this technology. In addition, we advocate for the establishment of robust oversight measures to ensure that the utilization of these advancements does not harm individuals or society.

\end{document}